\documentclass[sigconf]{acmart}

\ifdefined\mathsfit
  \let\WWWsavedmathsfit\mathsfit
  \let\mathsfit\relax
\fi
\usepackage{amsmath,amsfonts,bm}

\def\eqref#1{equation~\ref{#1}}
\def\1{\bm{1}}

\DeclareMathAlphabet{\mathsfit}{\encodingdefault}{\sfdefault}{m}{sl}
\SetMathAlphabet{\mathsfit}{bold}{\encodingdefault}{\sfdefault}{bx}{n}

\ifdefined\WWWsavedmathsfit
  \let\mathsfit\WWWsavedmathsfit
\fi

\setcopyright{none}
\renewcommand\footnotetextcopyrightpermission[1]{}  % no conference/copyright box on page 1 (submission)
\acmConference[Conference acronym 'XX]{Make sure to enter the correct conference title from your rights confirmation email}{June 03--05, 2018}{Woodstock, NY}
\acmYear{2018}
\copyrightyear{2018}
\acmDOI{}
\acmISBN{}

\usepackage{booktabs}
\usepackage{multirow}
\usepackage[table]{xcolor}
\definecolor{domainblue}{RGB}{232,242,252}
\usepackage{enumitem}
\usepackage{tabularx}
\usepackage{tcolorbox}
\definecolor{promptnavy}{RGB}{225,225,225}
\definecolor{promptbackground}{RGB}{255,255,255}
\usepackage{pifont}
\usepackage{pdfrender}
\newcommand{\graycross}{\textcolor{gray}{\ding{55}}}
\renewcommand{\graycross}{
  \textpdfrender{
    TextRenderingMode=FillStroke,
    FillColor=gray,
    StrokeColor=white,
    LineWidth=0.20pt
  }{\ding{55}}
}

\title{Circe: Training-Free LLM Self-Evaluation \\ via Component Interaction Consistency}
\author{Shuyi Miao}
\affiliation{%
  \institution{Beihang University}
  \city{Beijing}
  \country{China}
}

\author{Zheng Li}
\affiliation{%
  \institution{Nanjing University of Science and Technology}
  \city{Nanjing}
  \country{China}
}

\author{Xu Xie}
\affiliation{%
  \institution{MicroBT Inc.}
  \country{China}
}

\author{Linrui Xu}
\affiliation{%
  \institution{Central South University}
  \city{Changsha}
  \country{China}
}

\author{Fei Shen}
\authornote{Corresponding author.}
\affiliation{%
  \institution{National University of Singapore}
  \city{Singapore}
  \country{Singapore}
}

\author{Xiaoyu Du}
\affiliation{%
  \institution{Nanjing University of Science and Technology}
  \city{Nanjing}
  \country{China}
}

\author{Tat-Seng Chua}
\affiliation{%
  \institution{National University of Singapore}
  \city{Singapore}
  \country{Singapore}
}

\usepackage{colortbl}
\begin{document}

\begin{abstract}
As large language models (LLMs) are increasingly deployed in real-world applications, evaluating the correctness of their responses is crucial for reliable use. However, such evaluation is challenging because reference answers are rarely available in practice. 
Existing reference-free methods mainly rely on output signals, which may assign high scores to incorrect responses and thus do not necessarily reflect correctness. 
This motivates exploring the model’s internal computation for complementary correctness signals. 
Recent studies exploit internal states, but they mainly analyze
hidden-state trajectories across layers, which record only the
combined updates of attention and feed-forward networks (FFNs) and overlook how these components interact.
We therefore propose Circe, a training-free self-evaluation method that assesses a response using both internal and external signals, without reference answers or repeated sampling.
For internal signals, we first introduce component interaction
consistency (CIC), which measures the directional alignment between attention updates in earlier layers and attention-induced FFN output changes in later layers.
We further incorporate representation geometry (RG) to characterize normalized displacement and angular changes along layerwise hidden-state trajectories.
For the external signal, we convert the mean negative log-likelihood of generated tokens into a bounded likelihood confidence (LC) score.
Finally, we fuse CIC, RG, and LC into a unified correctness score.
Experiments on five LLMs and six benchmarks across four domains show that our method distinguishes correct from incorrect responses more effectively than existing baselines, and generalizes to multilingual, multimodal, and open-world settings. Our code is available at \url{https://github.com/laziestrp/CIRCE}.

\end{abstract}

\begin{CCSXML}
<ccs2012>
   <concept>
       <concept_id>10002944.10011123.10011130</concept_id>
       <concept_desc>General and reference~Evaluation</concept_desc>
       <concept_significance>500</concept_significance>
   </concept>
   <concept>
       <concept_id>10010147.10010178</concept_id>
       <concept_desc>Computing methodologies~Artificial intelligence</concept_desc>
       <concept_significance>300</concept_significance>
   </concept>
</ccs2012>
\end{CCSXML}

\ccsdesc[500]{General and reference~Evaluation}
\ccsdesc[300]{Computing methodologies~Artificial intelligence}

\keywords{Large Language Models, Self-Evaluation, Response Correctness, Latent Knowledge, Interpretability}

\maketitle

\section{Introduction}

\begin{figure}[!t]
  \centering
  \includegraphics[width=\linewidth]
  {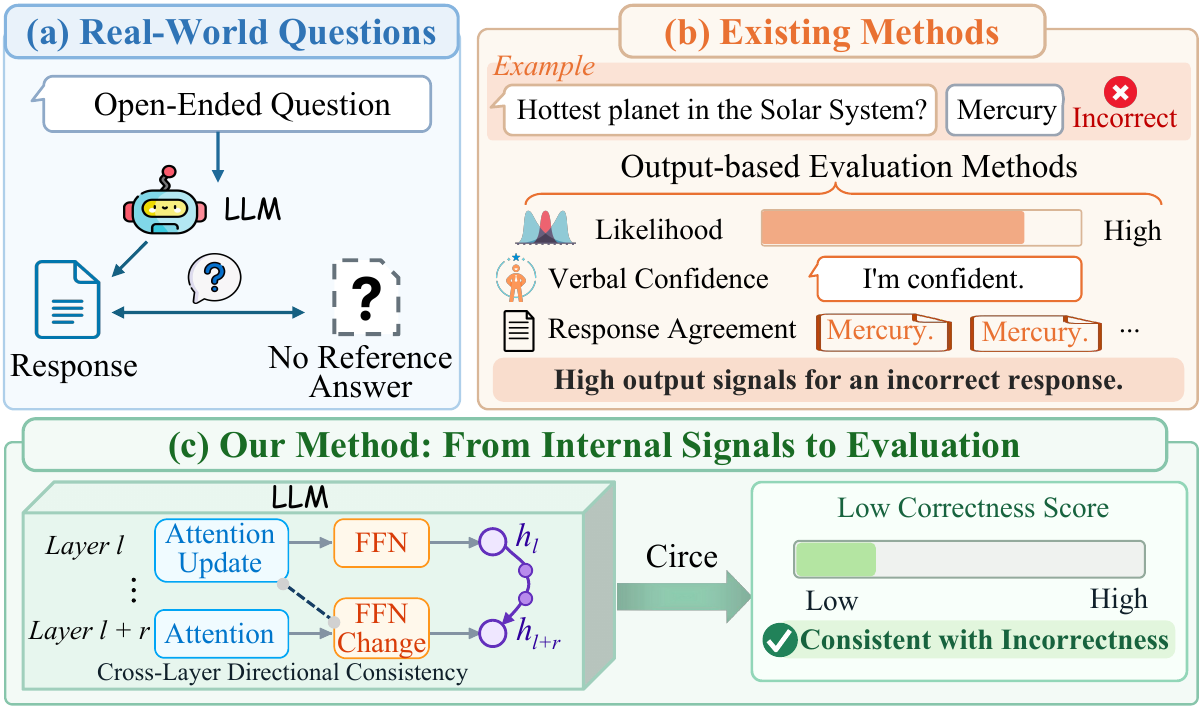}
  \vspace{-0.6cm}
\caption{\textbf{Evaluating LLM response correctness with internal signals.}
(a) Real-world questions often lack reference answers.
(b) Output likelihood, verbal confidence, and response agreement can remain high for incorrect answers.
(c) Circe incorporates internal computational signals, yielding a low correctness score for the illustrated incorrect answer.}
  \vspace{-0.6cm}
  \label{fig:motivation}
\end{figure}

Large language models (LLMs) have demonstrated strong capabilities in mathematical reasoning~\citep{ahn2024large}, knowledge question answering~\citep{wang2025bring}, language understanding~\citep{kuang2025natural,luo2026know,miao2026who}, and code generation~\citep{crupi2025effectiveness,miao2026contractlens}.
However, their responses can be fluent and plausible while still being incorrect, making correctness evaluation important for reliable deployment.
Reference answers provide a basis for verification, but are often unavailable for questions encountered in real-world applications, as illustrated in Figure~\ref{fig:motivation}(a).
This motivates reference-free evaluation methods that estimate response correctness from information available during or after generation.

Existing methods for reference-free correctness evaluation~\citep{ito2025reference} commonly derive signals from model outputs.
These include verbal confidence elicited through prompting~\citep{lin2022teaching,tian2023just}, probability-based measures such as entropy~\citep{Entropy} and perplexity~\citep{Perplexity}, and agreement across responses obtained through prompt perturbation~\citep{gao2024spuq} or repeated sampling~\citep{wang2023selfconsistency}.
However, these signals do not necessarily reflect response correctness: an incorrect answer may receive high probability, be accompanied by a confident self-assessment, or recur consistently across generations, as illustrated in Figure~\ref{fig:motivation}(b).
Methods requiring multiple responses also introduce additional inference costs.
These limitations motivate us to investigate internal computational signals that can complement output-level evidence.

Recent studies have explored correctness signals in internal representations~\citep{chen2024inside}, including layerwise hidden-state trajectories~\citep{wang2025coe}.
Such trajectories characterize the overall evolution of representations but leave the interactions between the underlying components implicit.
Within a Transformer block, attention aggregates contextual information, and the FFN transforms the updated representation.
Their contributions are combined in the residual stream, so similar aggregate changes can arise from different patterns of component interaction.
This raises a complementary question: can the relationship between attention updates and subsequent FFN responses provide additional evidence of response correctness?
We examine this relationship and observe that correct responses tend to exhibit higher directional alignment between earlier attention updates and attention-induced FFN output changes in later layers.
This observation motivates an interaction-based correctness signal alongside aggregate representation trajectories.

Building on this observation, we propose Circe, a training-free self-evaluation metric that combines component interactions, representation geometry, and output confidence.
Specifically, we first introduce \emph{Component Interaction Consistency (CIC)} to quantify the cross-layer alignment described above.
At each block, we compute the difference between the FFN outputs with and without the local attention update, and compare its direction with attention updates from earlier blocks.
Second, we incorporate \emph{Representation Geometry (RG)} to characterize normalized displacement and angular changes along layerwise hidden-state trajectories, providing complementary evidence from overall representation evolution.
Finally, \emph{Likelihood Confidence (LC)} transforms the mean negative log-likelihood of generated tokens into a bounded score that captures output-level confidence.
Circe first combines CIC and RG into an internal score, which is then fused with LC to produce a unified response-level correctness score, as illustrated in Figure~\ref{fig:motivation}(c).
The domain-specific fusion weights are selected on development data and fixed at test time.
Circe evaluates each test response using internal activations and token probabilities, without reference answers or repeated response sampling.
Our main contributions are as follows:
\begin{itemize}[leftmargin=*, topsep=4pt, partopsep=0pt]
    \item We identify cross-layer directional consistency between earlier attention updates and attention-induced FFN output changes in later layers as a correctness-related signal, providing component-level evidence that complements aggregate representation trajectories.

    \item We propose Circe (\underline{C}omponent \underline{I}nteraction, \underline{R}epresentation, and \underline{C}onfidence for \underline{E}valuation), a training-free self-evaluation metric that integrates component interactions, representation geometry, and output confidence to score individual responses without reference answers at inference time or repeated sampling.

    \item Extensive experiments on five LLMs and six benchmarks spanning four domains demonstrate competitive performance in response correctness evaluation, with ablation studies supporting the complementary roles of the three signals.
    Additional evaluations examine its applicability to multilingual and multimodal tasks and mixed benchmark distributions.
\end{itemize}

\section{Related Work}
Existing approaches to evaluating LLM response correctness can be broadly divided according to whether they rely on reference answers. Supervised methods learn to evaluate correctness from responses annotated against reference answers, whereas reference-free methods derive correctness signals without such references. 

\noindent\textbf{Supervised Correctness Evaluation.}
These methods learn to evaluate correctness from responses labelled as correct or incorrect according to reference answers. Hidden-state probing methods~\citep{azaria2023internal,orgad2025llms} train lightweight classifiers on internal representations, while external verifiers~\citep{cobbe2021gsm8k,lightman2024let} are trained with outcome-level or process-level annotations. These studies show that model representations encode information about response correctness beyond output likelihood or agreement. However, their reliance on reference-based annotations limits their applicability when reference answers are unavailable, and their performance may degrade under distribution shift between training and evaluation. These limitations motivate reference-free approaches to correctness evaluation.

\noindent\textbf{Reference-free Evaluation from Model Outputs.}
These methods evaluate response correctness using generated text and output distributions. 
Probability-based methods~\citep{Entropy,Perplexity} compute confidence measures such as negative log-likelihood, perplexity, and predictive entropy from token probabilities during generation. 
Verbal-confidence methods~\citep{lin2022teaching,tian2023just,kadavath2022} elicit an explicit self-assessment from the model, such as a stated confidence level or the probability it assigns to the correctness of its own answer. 
Another line of work estimates confidence from consistency across multiple responses. 
Self-Consistency~\citep{wang2023selfconsistency} derives confidence from answer agreement, Semantic Entropy~\citep{kuhn2023semantic} quantifies uncertainty over semantically equivalent responses, SelfCheckGPT~\citep{manakul2023selfcheckgpt} checks individual statements against sampled responses, and SPUQ~\citep{gao2024spuq} aggregates outputs under prompt and temperature perturbations. 
Although these require no reference answers, responses that are probable, confident, or mutually consistent can still be incorrect, and methods that require multiple generations incur additional inference cost.

\noindent\textbf{Reference-free Evaluation from Internal States.}
To exploit the correctness information encoded in internal representations without reference answers, recent work derives correctness scores directly from hidden states. EigenScore~\citep{chen2024inside} measures semantic consistency through the eigenvalues of the covariance matrix of hidden representations, but still requires multiple generations. CoE~\citep{wang2025coe} treats the hidden representations across layers as a Chain-of-Embedding and evaluates correctness from the magnitude and angular changes between adjacent layers, requiring neither reference answers nor repeated sampling. However, such trajectory statistics summarize the aggregate evolution of representations, leaving the interaction between attention and FFNs within each layer implicit. In contrast, our method explicitly measures the directional consistency between attention updates and the FFN output changes they induce, and combines it with representation geometry and output confidence to evaluate the response.

\begin{figure*}[!t]
  \centering
  \includegraphics[width=\linewidth]
  {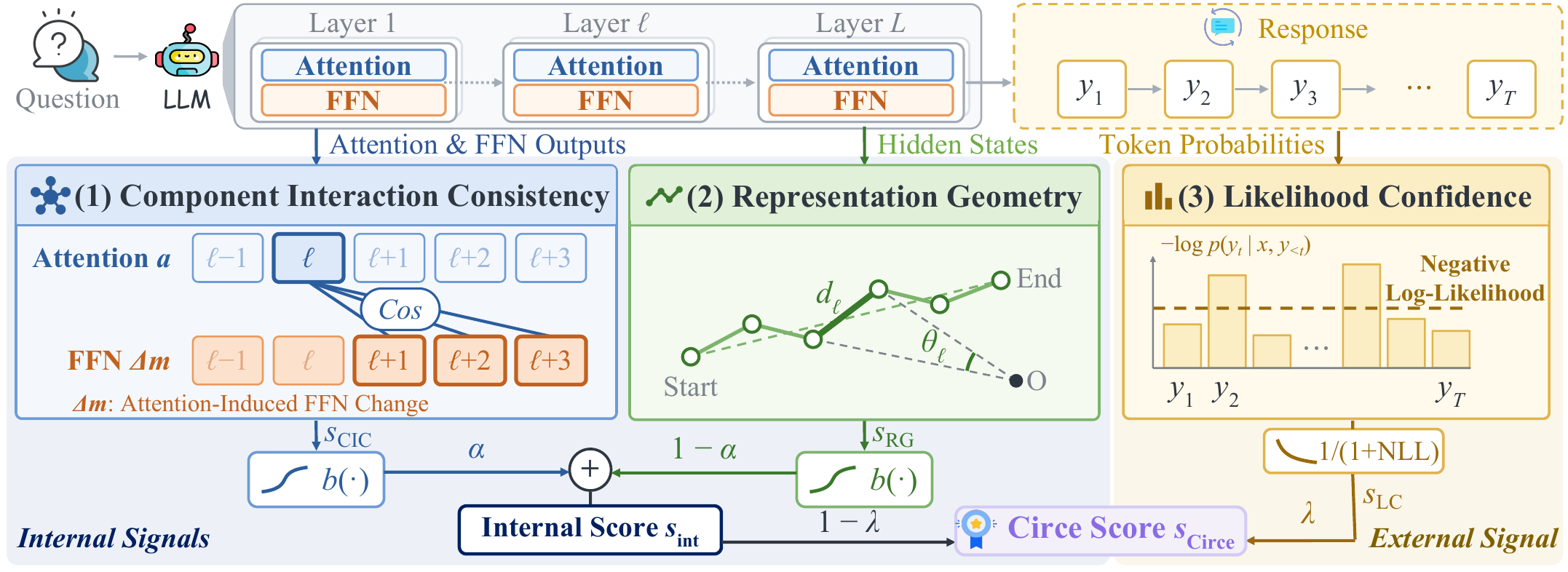}
  \vspace{-0.7cm}
    \caption{Overview of Circe. (1) Component interaction consistency: cross-layer directional alignment between attention updates and attention-induced FFN changes. (2) Representation geometry: normalized displacement and angular change along the layerwise hidden state trajectory. (3) Likelihood confidence: a bounded transformation of the mean NLL of the generated tokens. Internal signals are fused into an internal score, which is combined with the output signal to yield the final Circe score.}
  \vspace{-0.2cm}
  \label{fig:framework}
\end{figure*}

\section{Method}
\label{sec:method}
\label{sec:cic}
\label{sec:score}

We estimate response correctness from internal and external
perspectives. The internal signals consist of CIC, which captures
component-level interactions across layers, and RG, which describes the overall evolution of residual representations. 
The external signal is LC, which reflects the probabilities assigned to generated tokens. 
As illustrated in Figure~\ref{fig:framework}, CIC and RG are first fused into an internal score, which is then combined with LC to obtain the final Circe score.

\subsection{Internal Evaluation Signals}
The internal signals examine the model from two complementary levels: CIC analyzes the directional consistency between internal components within each layer, whereas RG analyzes the geometry of the hidden states that each layer finally outputs.

\begin{figure*}[!t]
  \centering
  \includegraphics[width=\linewidth]
  {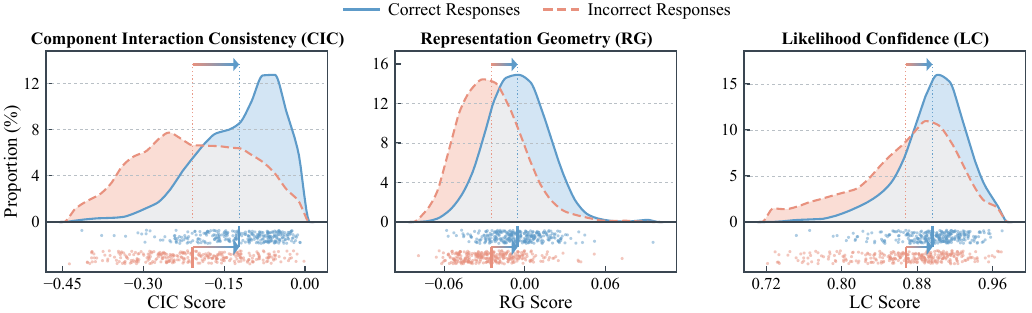}
  \vspace{-0.6cm}
    \caption{\textbf{Correct responses have higher mean CIC, RG, and LC scores.}
    Curves show smoothed score distributions for Qwen2-7B-Instruct on MATH, and dots below the curves represent individual responses.
Vertical dotted lines and short bars mark the class means, with arrows pointing from the incorrect-response mean to the correct-response mean.
Grey shading indicates overlap between the distributions of the two classes.
    }
  \vspace{-0.3cm}
  \label{fig:3view}
\end{figure*}

\noindent\textbf{Component Interaction Consistency Score.}
Attention and FFNs jointly shape information processing across
Transformer layers. We examine whether attention updates in
earlier blocks align with attention-related changes in FFN
outputs in later blocks. We think that stronger alignment
reflects more coherent information processing and is associated
with a higher likelihood of a correct response.
To characterize this interaction, we first quantify how an
attention update changes the FFN output within the same block.
Consider a Transformer with $L$ blocks and a response of $T$
tokens. At token position $t\in\{1,\ldots,T\}$ and block
$\ell\in\{1,\ldots,L\}$, let $h_{t,\ell-1}$ denote the incoming
residual state. Let $a_{t,\ell}$ denote the attention update
added to this state, and $F_\ell$ the FFN computation, including
its input normalization. We compare the FFN output with and
without the current attention update:
\begin{equation}
    \Delta m_{t,\ell}
    = F_\ell(h_{t,\ell-1}+a_{t,\ell})
      -F_\ell(h_{t,\ell-1}).
\end{equation}
Both evaluations use the same incoming residual state and
model parameters. This difference captures the change in FFN
output induced by adding the current attention update. Its
direction therefore characterizes how attention modifies the
FFN contribution to the residual stream within that block.
To obtain response-level vectors for each block, we average
the attention updates and attention-induced FFN output changes
over all response tokens:
\begin{equation}
\begin{aligned}
    \bar a_{\ell}
    &= \frac{1}{T}
       \sum_{t=1}^{T} a_{t,\ell},\\
    \overline{\Delta m}_{\ell}
    &= \frac{1}{T}
       \sum_{t=1}^{T} \Delta m_{t,\ell}.
\end{aligned}
\end{equation}
% 
% \noindent\textbf{Component Interaction Consistency Score.}
To characterize the persistence of directional alignment
across layers, we compare earlier attention updates with
locally defined attention-induced FFN output changes in
subsequent blocks. We aggregate these cosine similarities to obtain the CIC score:
\begin{equation}
    s_{\mathrm{CIC}}
    = \frac{1}{R}
      \sum_{r=1}^{R}
      \frac{1}{L-r}
      \sum_{\ell=1}^{L-r}
      \cos\!\left(
          \bar a_{\ell},
          \overline{\Delta m}_{\ell+r}
      \right),
\end{equation}
where $r$ denotes the layer separation, $R$ its maximum value
($1\leq R<L$), and $\cos(\cdot,\cdot)$ cosine similarity
between nonzero vectors in the shared residual-stream space.
Higher $s_{\mathrm{CIC}}$ indicates stronger average alignment between earlier attention directions and attention-induced FFN output changes across subsequent layers, as shown by the CIC scores for Qwen2-7B-Instruct on MATH in Figure~\ref{fig:3view}.

\noindent\textbf{Representation Geometry Score.}
Following ~\citep{wang2025coe},
we construct a representation trajectory by averaging the
residual states over all response tokens at each layer:
\begin{equation}
    \bar h_{\ell}
    = \frac{1}{T}
      \sum_{t=1}^{T} h_{t,\ell},
    \qquad \ell\in\{0,\ldots,L\}.
\end{equation}
The sequence $(\bar h_0,\ldots,\bar h_L)$ describes how the
response-level representation evolves across layers.
$h_{t,0}$ is the residual input to the first block,
and $h_{t,L}$ is the output of the final block.

We characterize each transition by its displacement and
angular change, normalized by the corresponding changes
between the initial and final states:
\begin{equation}
    d_{\ell}
    = \frac{\|\bar h_{\ell+1}-\bar h_{\ell}\|_2}
           {\|\bar h_L-\bar h_0\|_2+\epsilon},
\end{equation}
\begin{equation}
    \theta_{\ell}
    = \frac{
        \arccos\!\bigl(
            \cos(\bar h_{\ell},\bar h_{\ell+1})
        \bigr)
      }{
        \arccos\!\bigl(
            \cos(\bar h_0,\bar h_L)
        \bigr)+\epsilon
      },
\end{equation}
where $\ell\in\{0,\ldots,L-1\}$, $\|\cdot\|_2$ denotes
the Euclidean norm, and $\epsilon>0$ ensures numerical
stability in the denominators.
Cosine values are clipped to $[-1,1]$ before applying
$\arccos$.
We combine the two quantities to obtain the RG score:
\begin{equation}
    s_{\mathrm{RG}}
    = \frac{1}{L}
      \sum_{\ell=0}^{L-1}
      \left(d_{\ell}-\theta_{\ell}\right).
\end{equation}
This score increases with normalized displacement and
decreases with normalized angular change, summarizing
representation evolution alongside the component consistency
measured by CIC.

%主表
\begin{table*}[!t]
\centering
\caption{\textbf{Evaluation performance across five models and four domains.}
Results are reported in AUROC (\%, $\uparrow$), FPR95 (\%, $\downarrow$), and AUPR (\%, $\uparrow$).
Average denotes the mean over the five models.
% Bold and underlined values indicate the best result and the best baseline result, respectively.
$\dagger$ denotes methods that require multiple stochastic inferences.}
\vspace{-0.3cm}
\label{tab:circe-four-domains}
\begingroup
\footnotesize
\setlength{\tabcolsep}{2.5pt}
\renewcommand{\arraystretch}{1.00}
\begin{tabular*}{\textwidth}{@{\extracolsep{\fill}}lcccccc@{}}
\toprule
\multirow{2}{*}{\textbf{Method}}
& \multicolumn{5}{c}{\textbf{Model}}
& \multirow{2}{*}{\textbf{Average}} \\
\cmidrule(lr){2-6}
& Qwen2-1.5B-Instruct\citep{yang2024qwen2}
& Qwen2-7B-Instruct\citep{yang2024qwen2}
& Llama3-8B-Instruct\citep{meta2024llama3}
& Mistral-7B-Instruct\citep{jiang2023mistral}
& Qwen2.5-14B-Instruct\citep{yang2024qwen25} & \\
\midrule
% Domain I
\rowcolor{domainblue}
\multicolumn{7}{c}{\textbf{Domain I: Mathematics}} \\
\midrule
\textit{Verbal Conf.}~\citep{kadavath2022} & 52.10 / 88.70 / 38.05 & 56.21 / 91.42 / 68.23 & 44.19 / 98.94 / 43.56 & 43.26 / 97.16 / 27.85 & 30.06 / 98.47 / 71.97 & 45.16 / 94.94 / 49.93 \\
\textit{PSA pipeline}$^{\dagger}$~\citep{xiong2024can} & 69.58 / 84.75 / 52.87 & 62.28 / 85.53 / 71.15 & 61.34 / 86.24 / 51.24 & \underline{75.68} / \underline{79.26} / \underline{44.98} & 66.57 / 77.37 / 87.69 & 67.09 / 82.63 / 61.59 \\
Max Prob.~\citep{hendrycks2017baseline} & 63.27 / 82.28 / 44.11 & 68.40 / 75.67 / 73.82 & 58.47 / 84.60 / 54.81 & 65.53 / 81.09 / 37.33 & 62.03 / 79.27 / 82.95 & 63.54 / 80.58 / 58.60 \\
Perplexity~\citep{Perplexity} & 63.78 / 82.07 / 44.48 & 68.83 / 73.09 / 74.23 & 58.32 / 84.71 / 54.72 & 66.37 / 79.40 / 37.93 & 62.47 / 79.11 / 83.21 & 63.95 / 79.68 / 58.91 \\
Entropy~\citep{Entropy} & 65.41 / 80.92 / 46.00 & 70.25 / 73.45 / 74.97 & 60.17 / 84.02 / 55.57 & 66.55 / 82.02 / 38.27 & 63.34 / 78.43 / 83.46 & 65.14 / 79.77 / 59.65 \\
Temp. Scaling~\citep{guo2017calibration} & 62.39 / 83.24 / 43.46 & 69.11 / 76.98 / 73.12 & 58.72 / 84.33 / 54.77 & 64.98 / 82.48 / 37.03 & 61.55 / 79.85 / 82.90 & 63.35 / 81.38 / 58.26 \\
Energy~\citep{liu2020energy} & \underline{74.80} / \underline{71.62} / \underline{61.11} & 56.10 / 88.52 / 70.76 & 53.26 / 92.64 / 50.32 & 49.07 / 96.57 / 26.94 & \underline{71.98} / 76.94 / \underline{88.50} & 61.04 / 85.26 / 59.53 \\
MC Dropout$^{\dagger}$~\citep{gal2016dropout} & 46.45 / 95.95 / 36.08 & 55.64 / 90.53 / 68.58 & 48.75 / 96.25 / 47.18 & 49.09 / 94.12 / 27.26 & 49.82 / 94.95 / 79.99 & 49.95 / 94.36 / 51.82 \\
LN-Entropy$^{\dagger}$~\citep{malinin2020uncertainty} & 70.92 / 78.06 / 54.45 & 66.63 / 71.09 / 73.84 & 61.95 / 85.61 / 56.63 & 67.71 / 80.16 / 39.25 & 66.28 / \underline{74.25} / 85.18 & 66.70 / \underline{77.83} / 61.87 \\
EigenScore$^{\dagger}$~\citep{chen2024inside} & 63.76 / 90.09 / 50.22 & 62.56 / 88.89 / 67.10 & 52.77 / 95.45 / 43.00 & 41.65 / 96.77 / 24.11 & 62.87 / 83.10 / 84.88 & 56.72 / 90.86 / 53.86 \\
CoE-R~\citep{wang2025coe} & 71.85 / 78.86 / 54.34 & \underline{75.70} / \underline{69.30} / \underline{83.37} & 73.19 / \underline{72.30} / 66.84 & 68.38 / 82.83 / 38.41 & 60.46 / 86.57 / 83.05 & 69.91 / 77.97 / 65.20 \\
CoE-C~\citep{wang2025coe} & 69.66 / 81.29 / 53.59 & 73.63 / 72.53 / 81.87 & \underline{74.03} / 77.52 / \underline{69.19} & 71.00 / 79.57 / 40.20 & 61.97 / 88.74 / 85.23 & \underline{70.06} / 79.93 / \underline{66.02} \\
\midrule
\textbf{Ours} & \textbf{79.04} / \textbf{63.63} / \textbf{65.48} & \textbf{77.68} / \textbf{64.91} / \textbf{84.69} & \textbf{79.21} / \textbf{62.86} / \textbf{73.72} & \textbf{75.84} / \textbf{66.18} / \textbf{47.01} & \textbf{75.40} / \textbf{63.17} / \textbf{90.56} & \textbf{77.43} / \textbf{64.15} / \textbf{72.29} \\
\midrule
\rowcolor{domainblue}
\multicolumn{7}{c}{\textbf{Domain II: Reasoning}} \\
\textit{Verbal Conf.}~\citep{kadavath2022} & 50.96 / 93.09 / 37.44 & 43.25 / 99.15 / 42.73 & 50.01 / 97.74 / 45.62 & 44.83 / 96.71 / 36.51 & 40.57 / 96.90 / 54.94 & 45.92 / 96.72 / 43.45 \\
\textit{PSA pipeline}$^{\dagger}$~\citep{xiong2024can} & 58.57 / 97.30 / 38.75 & 48.98 / 97.59 / 49.53 & 51.25 / 94.76 / 47.08 & 46.07 / 97.30 / 38.84 & 60.46 / 91.43 / 65.24 & 53.07 / 95.68 / 47.89 \\
Max Prob.~\citep{hendrycks2017baseline} & 63.84 / 83.03 / 39.63 & 54.57 / 91.77 / 54.21 & 48.14 / 95.10 / 44.43 & 53.90 / 91.50 / 39.53 & 60.23 / 85.08 / 63.18 & 56.14 / 89.30 / 48.20 \\
Perplexity~\citep{Perplexity} & 64.05 / 84.26 / 39.91 & 55.39 / 93.33 / 54.78 & 48.40 / 95.62 / 44.53 & 54.70 / 90.84 / 40.02 & 60.74 / 84.75 / 63.54 & 56.66 / 89.76 / 48.56 \\
Entropy~\citep{Entropy} & 63.84 / 79.51 / 39.71 & 55.97 / 92.69 / 54.80 & 48.56 / 95.55 / 44.64 & 54.83 / 90.04 / 40.12 & 60.92 / 83.51 / 63.50 & 56.82 / 88.26 / 48.55 \\
Temp. Scaling~\citep{guo2017calibration} & 63.55 / 83.40 / 39.75 & 54.06 / 94.04 / 53.87 & 48.12 / 95.47 / 44.39 & 53.43 / 92.23 / 39.23 & 60.11 / 85.27 / 63.28 & 55.85 / 90.08 / 48.10 \\
Energy~\citep{liu2020energy} & 61.84 / 77.87 / 41.74 & 47.52 / 95.16 / 50.07 & 43.69 / 97.02 / 41.29 & 48.34 / 95.61 / 36.84 & \underline{67.55} / \underline{82.72} / \underline{70.76} & 53.79 / 89.68 / 48.14 \\
MC Dropout$^{\dagger}$~\citep{gal2016dropout} & 45.51 / 95.44 / 31.62 & 50.14 / 97.17 / 49.86 & 46.21 / 96.60 / 42.07 & 51.29 / 95.09 / 37.44 & 49.33 / 94.05 / 57.89 & 48.50 / 95.67 / 43.78 \\
LN-Entropy$^{\dagger}$~\citep{malinin2020uncertainty} & 67.32 / \underline{74.00} / 43.75 & 56.07 / 94.25 / 54.26 & 52.88 / 95.36 / 40.92 & 59.74 / 92.26 / 44.77 & 61.24 / 87.15 / 64.14 & 59.45 / 88.60 / 49.57 \\
EigenScore$^{\dagger}$~\citep{chen2024inside} & 61.53 / 88.55 / 40.38 & 53.39 / 94.11 / 53.39 & 53.58 / 95.94 / 45.80 & 60.70 / 91.12 / 48.29 & 62.32 / 87.78 / 67.71 & 58.30 / 91.50 / 51.11 \\
CoE-R~\citep{wang2025coe} & 69.30 / 80.81 / 48.80 & \underline{67.18} / \underline{83.75} / \underline{64.88} & \underline{63.08} / \underline{90.10} / \underline{54.96} & \underline{72.22} / \underline{79.18} / \underline{58.09} & 62.27 / 86.72 / 67.70 & \underline{66.81} / \underline{84.11} / \underline{58.89} \\
CoE-C~\citep{wang2025coe} & \underline{69.88} / 76.38 / \underline{48.81} & 62.36 / 90.33 / 63.76 & 55.85 / 90.12 / 50.48 & 69.39 / 82.93 / 57.59 & 64.22 / 84.68 / 69.41 & 64.34 / 84.89 / 58.01 \\
\midrule
\textbf{Ours} & \textbf{74.62} / \textbf{68.25} / \textbf{52.98} & \textbf{67.94} / \textbf{80.45} / \textbf{66.09} & \textbf{63.48} / \textbf{86.23} / \textbf{55.12} & \textbf{74.22} / \textbf{74.84} / \textbf{59.35} & \textbf{69.46} / \textbf{79.73} / \textbf{74.35} & \textbf{69.94} / \textbf{77.90} / \textbf{61.58} \\
\midrule
\rowcolor{domainblue}
\multicolumn{7}{c}{\textbf{Domain III: Knowledge}} \\
\midrule
\textit{Verbal Conf.}~\citep{kadavath2022} & 57.49 / 97.28 / 56.95 & 51.25 / 96.61 / 49.62 & 46.37 / 99.56 / 59.34 & 53.52 / 92.78 / 51.75 & 35.38 / 95.79 / 72.13 & 48.80 / 96.40 / 57.96 \\
\textit{PSA pipeline}$^{\dagger}$~\citep{xiong2024can} & 49.79 / 94.68 / 50.22 & \textbf{\underline{60.25}} / 88.81 / 70.10 & 54.95 / \underline{86.34} / 69.64 & 56.67 / 92.43 / 55.45 & 63.67 / 87.72 / 87.14 & 57.07 / 90.00 / 66.51 \\
Max Prob.~\citep{hendrycks2017baseline} & 60.54 / 90.83 / 57.47 & 57.09 / 95.31 / 71.76 & 49.92 / 93.99 / 66.13 & 53.15 / 96.28 / 52.42 & 67.18 / 85.88 / 87.99 & 57.58 / 92.46 / 67.15 \\
Perplexity~\citep{Perplexity} & \underline{61.13} / 90.22 / 58.04 & 57.26 / 94.79 / \underline{72.04} & 50.50 / 92.90 / 66.59 & 53.41 / 95.95 / 52.95 & 67.50 / 85.66 / 88.15 & 57.96 / 91.90 / 67.55 \\
Entropy~\citep{Entropy} & 60.77 / 90.34 / 57.29 & 57.80 / 93.75 / 71.84 & 50.12 / 92.35 / 66.25 & 54.39 / 94.59 / 53.42 & 68.08 / 84.81 / 88.36 & 58.23 / 91.17 / 67.43 \\
Temp. Scaling~\citep{guo2017calibration} & 60.30 / 90.81 / 58.13 & 56.81 / 93.75 / 71.40 & 49.52 / 93.99 / 65.86 & 53.18 / 96.96 / 52.14 & 66.78 / 86.12 / 87.86 & 57.32 / 92.33 / 67.08 \\
Energy~\citep{liu2020energy} & 56.98 / 92.98 / 53.30 & 48.73 / 97.72 / 64.79 & 46.90 / 95.77 / 61.74 & 46.58 / 99.08 / 45.73 & \underline{70.85} / 84.25 / \underline{90.20} & 54.01 / 93.96 / 63.15 \\
MC Dropout$^{\dagger}$~\citep{gal2016dropout} & 52.62 / 94.29 / 51.82 & 55.64 / 95.88 / 68.78 & 49.87 / 93.16 / 65.98 & 50.24 / 98.43 / 50.69 & 48.25 / 95.40 / 79.07 & 51.32 / 95.43 / 63.27 \\
LN-Entropy$^{\dagger}$~\citep{malinin2020uncertainty} & 60.63 / \underline{90.09} / \underline{59.22} & 58.01 / 92.68 / 71.99 & 55.36 / 91.14 / 68.90 & 60.42 / 90.36 / 57.63 & 69.68 / \underline{83.79} / 89.04 & \underline{60.82} / 89.61 / \underline{69.36} \\
EigenScore$^{\dagger}$~\citep{chen2024inside} & 50.82 / 95.04 / 50.80 & 50.36 / 96.35 / 67.53 & 58.30 / 91.80 / 69.73 & 61.68 / 89.53 / \underline{62.75} & 64.15 / 86.51 / 86.69 & 57.06 / 91.85 / 67.50 \\
CoE-R~\citep{wang2025coe} & 49.33 / 93.36 / 50.45 & 57.98 / \underline{87.50} / 71.54 & \underline{60.19} / 88.13 / \underline{72.19} & 62.88 / \underline{88.16} / 61.04 & 64.57 / 90.80 / 87.60 & 58.99 / \underline{89.59} / 68.56 \\
CoE-C~\citep{wang2025coe} & 49.57 / 93.81 / 50.83 & 57.56 / 88.14 / 71.45 & 55.68 / 89.87 / 68.27 & \underline{63.73} / 88.73 / 62.53 & 66.03 / 92.11 / 88.67 & 58.51 / 90.53 / 68.35 \\
\midrule
\textbf{Ours} & \textbf{63.70} / \textbf{87.00} / \textbf{61.23} & 59.79 / \textbf{86.22} / \textbf{75.14} & \textbf{66.60} / \textbf{85.37} / \textbf{77.44} & \textbf{67.20} / \textbf{86.24} / \textbf{65.26} & \textbf{72.38} / \textbf{83.29} / \textbf{91.01} & \textbf{65.93} / \textbf{85.62} / \textbf{74.02} \\
\midrule
\rowcolor{domainblue}
\multicolumn{7}{c}{\textbf{Domain IV: Understanding}} \\
\midrule
\textit{Verbal Conf.}~\citep{kadavath2022} & 49.83 / 96.83 / 71.11 & 51.18 / 95.67 / 79.65 & 47.42 / 99.10 / 83.64 & 50.21 / 97.33 / 45.24 & 40.23 / 83.33 / 93.06 & 47.77 / 94.45 / 74.54 \\
\textit{PSA pipeline}$^{\dagger}$~\citep{xiong2024can} & 57.31 / 91.55 / 74.44 & 62.07 / 93.20 / 92.15 & 52.07 / 89.64 / 89.37 & 54.19 / 94.88 / 50.82 & 63.48 / 80.95 / 97.04 & 57.82 / 90.04 / 80.76 \\
Max Prob.~\citep{hendrycks2017baseline} & 60.02 / \textbf{\underline{87.32}} / 74.38 & 60.52 / 88.35 / 91.80 & 56.64 / 95.19 / 90.42 & 60.82 / 92.32 / 56.70 & 64.96 / 83.33 / 97.03 & 60.59 / 89.30 / 82.07 \\
Perplexity~\citep{Perplexity} & 60.78 / 92.25 / 75.46 & 60.93 / 87.38 / 91.80 & 56.64 / 94.23 / 90.44 & 61.38 / 91.92 / 57.50 & 65.34 / 83.33 / 96.99 & 61.01 / 89.82 / 82.44 \\
Entropy~\citep{Entropy} & 59.80 / 88.03 / 73.15 & 62.65 / 86.41 / 92.13 & 56.78 / 92.31 / 90.43 & 61.87 / 91.11 / 58.11 & 66.40 / 85.71 / 97.15 & 61.50 / 88.71 / 82.19 \\
Temp. Scaling~\citep{guo2017calibration} & 60.30 / 89.44 / \underline{76.55} & 60.45 / 88.35 / 91.84 & 56.63 / 95.19 / 90.38 & 60.35 / 92.12 / 56.07 & 64.10 / 83.33 / 96.94 & 60.37 / 89.69 / 82.36 \\
Energy~\citep{liu2020energy} & 56.43 / 91.90 / 70.16 & 53.26 / 97.38 / 81.67 & 44.78 / 98.06 / 80.14 & 50.78 / 95.64 / 50.22 & 68.98 / \underline{76.19} / 97.14 & 54.85 / 91.83 / 75.87 \\
MC Dropout$^{\dagger}$~\citep{gal2016dropout} & 50.49 / 94.01 / 67.82 & 58.23 / 91.42 / 87.09 & 51.20 / 97.89 / 85.68 & 53.97 / 95.36 / 52.55 & 43.48 / 100.00 / 94.03 & 51.47 / 95.74 / 77.43 \\
LN-Entropy$^{\dagger}$~\citep{malinin2020uncertainty} & \underline{61.76} / 91.55 / \underline{76.55} & 59.65 / 90.50 / 88.61 & 55.42 / 92.68 / 88.79 & 58.71 / 89.13 / 49.54 & \textbf{\underline{71.31}} / 80.95 / \textbf{\underline{97.72}} & 61.37 / 88.96 / 80.24 \\
EigenScore$^{\dagger}$~\citep{chen2024inside} & 52.34 / 94.72 / 70.27 & 63.59 / 90.29 / 92.79 & 56.32 / 91.35 / 90.19 & 60.32 / 90.71 / 53.58 & 59.25 / 95.24 / 96.82 & 58.36 / 92.46 / 80.73 \\
CoE-R~\citep{wang2025coe} & 53.61 / 92.96 / 72.84 & \underline{73.55} / \underline{72.92} / \underline{95.24} & \underline{68.55} / \underline{86.36} / \underline{94.57} & 62.99 / \textbf{\underline{82.09}} / 85.65 & 63.98 / 85.71 / 96.46 & \underline{64.54} / \underline{84.01} / \underline{88.95} \\
CoE-C~\citep{wang2025coe} & 53.88 / 92.96 / 73.15 & 72.47 / 78.12 / 95.00 & 61.09 / 90.91 / 92.99 & \underline{63.36} / 88.56 / \underline{85.87} & 65.48 / 85.71 / 96.74 & 63.25 / 87.25 / 88.75 \\
\midrule
\textbf{Ours} & \textbf{63.42} / 87.68 / \textbf{79.57} & \textbf{75.10} / \textbf{66.67} / \textbf{95.27} & \textbf{73.61} / \textbf{78.41} / \textbf{95.62} & \textbf{66.89} / 83.08 / \textbf{87.18} & 70.21 / \textbf{66.67} / 97.16 & \textbf{69.85} / \textbf{76.50} / \textbf{90.96} \\
\bottomrule
\end{tabular*}
\endgroup
\end{table*}

\subsection{External Evaluation Signal}

To complement the two internal signals, we examine the
probabilities assigned to the tokens of the generated response
$y=(y_1,\ldots,y_T)$ for a prompt $x$, and summarize them into the LC score.
Motivated by the likelihood-based uncertainty term, we compute the mean token negative log-likelihood (NLL):
\begin{equation}
    \mathrm{NLL}
    = -\frac{1}{T}
      \sum_{t=1}^{T}
      \log p(y_t\mid x,y_{<t}),
\end{equation}
where $p(y_t\mid x,y_{<t})$ is the probability assigned to
the generated token $y_t$ given the prompt and preceding
response tokens.
Lower NLL indicates a higher geometric mean of the probabilities assigned to the generated tokens.
We transform NLL into the LC score:
\begin{equation}
    s_{\mathrm{LC}}
    = \frac{1}{1+\mathrm{NLL}}.
\end{equation}
Higher $s_{\mathrm{LC}}$ therefore indicates greater
likelihood-based confidence in the generated response.
For Qwen2-7B-Instruct on MATH, correct responses have higher
mean RG and LC scores than incorrect responses, as shown
in Figure~\ref{fig:3view}.

% 组件消融
\begin{table*}[!t]
\centering
\caption{\textbf{Component ablation of our method across five models.} AUROC (\%, $\uparrow$) is averaged over six benchmarks for each model. ``only'' uses a single component, and ``w/o'' removes the component from the full method.}
\vspace{-0.3cm}
\label{tab:circe_ablation}
\small
\renewcommand{\arraystretch}{1.12}

\newcommand{\circeAblationRows}{%
\toprule
& \multicolumn{3}{c}{Components}
& \multicolumn{6}{c}{AUROC} \\
\cmidrule(lr){2-4}
\cmidrule(lr){5-10}
Variant
& CIC & RG & LC
& Qwen2-1.5B-Instruct\citep{yang2024qwen2}
& Qwen2-7B-Instruct\citep{yang2024qwen2}
& Llama3-8B-Instruct\citep{meta2024llama3}
& Mistral-7B-Instruct\citep{jiang2023mistral}
& Qwen2.5-14B-Instruct\citep{yang2024qwen25}
& Average \\
\midrule

CIC only
& $\checkmark$ & \graycross & \graycross
& 59.22 & 68.72 & 66.21 & 66.33 & 69.01 & 65.90 \\

LC only
& \graycross & \graycross & $\checkmark$
& 62.93 & 60.28 & 53.77 & 58.19 & 62.44 & 59.52 \\

RG only
& \graycross & $\checkmark$ & \graycross
& 64.21 & 69.55 & 66.88 & 67.84 & 62.33 & 66.16 \\

\midrule

w/o LC
& $\checkmark$ & $\checkmark$ & \graycross
& 65.32 & 70.36 & 68.88 & 70.14 & 69.68 & 68.87 \\

w/o CIC
& \graycross & $\checkmark$ & $\checkmark$
& 69.38 & 70.43 & 69.46 & 70.43 & 68.83 & 69.71 \\

w/o RG
& $\checkmark$ & \graycross & $\checkmark$
& 70.64 & 69.34 & 67.38 & 68.65 & 70.02 & 69.21 \\

\midrule

\rowcolor{domainblue}
\textbf{Ours}
& $\checkmark$ & $\checkmark$ & $\checkmark$
& \textbf{72.41} & \textbf{71.02} & \textbf{70.93}
& \textbf{72.37} & \textbf{72.05} & \textbf{71.76} \\

\bottomrule
}

\setlength{\tabcolsep}{6pt}
\sbox0{%
  \begin{tabular}{lccc*{6}{c}}
    \circeAblationRows
  \end{tabular}%
}

\setlength{\tabcolsep}{
  \dimexpr 6pt+(\textwidth-\wd0)/20\relax
}

\begin{tabular}{lccc*{6}{c}}
  \circeAblationRows
\end{tabular}

\end{table*}

\subsection{Constructing the Circe Score}
We combine component interaction consistency, representation geometry, and output confidence into the Circe score.
Before fusion, we map the two internal scores to $(0,1)$:
\begin{equation}
    b(z)
    = \frac{1}{2}
      \left(1+\frac{z}{1+|z|}\right),
\end{equation}
where $z$ is a scalar score.
This strictly increasing transformation preserves the
ordering of each internal score.
We first combine the transformed CIC and RG scores:
\begin{equation}
    s_{\mathrm{int}}
    = \alpha b(s_{\mathrm{CIC}})
      +(1-\alpha)b(s_{\mathrm{RG}}),
\end{equation}
where $\alpha\in[0,1]$ balances the contributions of the
two internal signals.
We then incorporate LC to obtain the final Circe score:
\begin{equation}
    s_{\mathrm{Circe}}
    = (1-\lambda)s_{\mathrm{int}}
      +\lambda s_{\mathrm{LC}},
\end{equation}
where $\lambda\in[0,1]$ controls the contribution of
output-level confidence.
Higher $s_{\mathrm{Circe}}$ indicates stronger combined
evidence of response correctness.
Appendix~\ref{app:theory} provides a theoretical analysis of the proposed scores and the rationale for their fusion.

\section{Experiments}

\subsection{Setup}

\noindent\textbf{Datasets.}
We evaluate the discriminative effectiveness of correctness metrics on six
public benchmarks, covering four different task types: mathematics, reasoning,
knowledge, and understanding.
Specifically, these include mathematics: GSM8K \citep{cobbe2021gsm8k} and MATH \citep{hendrycks2021math}, reasoning: CommonsenseQA \citep{talmor2019commonsenseqa} and TheoremQA \citep{chen2023theoremqa}, knowledge: MMLU \citep{hendrycks2021mmlu}, and understanding: Belebele \citep{bandarkar-etal-2024-belebele}.

\noindent\textbf{Models.}
We select five open-source models with diverse architectures and training origins: Qwen2-1.5B-Instruct \citep{yang2024qwen2}, Qwen2-7B-Instruct \citep{yang2024qwen2}, Llama3-8B-Instruct \citep{meta2024llama3}, Mistral-7B-Instruct \citep{jiang2023mistral}, and Qwen2.5-14B-Instruct \citep{yang2024qwen25}.
% 补充附录
The instructions used by all models to generate responses are listed in Appendix~\ref{app:prompts}.
% We use the official instruction-tuned versions of each model together with their corresponding chat templates, and adopt a unified prompt format for all tasks.

\noindent\textbf{Baselines.}
We compare against 12 reference-free evaluation baselines, grouped by the source of their correctness signals.
Output-based methods derive correctness scores from generated text, output probabilities, or logits. Among single-pass methods, we select P(True)~\citep{kadavath2022} for verbal confidence (verbal conf.) because of its versatility. Maximum softmax probability (max prob.)~\citep{hendrycks2017baseline}, perplexity~\citep{Perplexity}, entropy~\citep{Entropy}, temperature scaling (temp. scaling)~\citep{guo2017calibration}, and energy~\citep{liu2020energy} compute confidence or uncertainty measures from output probabilities or logits. Among methods requiring multiple inferences, the PSA pipeline~\citep{xiong2024can} aggregates confidence over prompted and sampled responses, Monte Carlo dropout (MC dropout)~\citep{gal2016dropout} aggregates predictions across stochastic forward passes, and length-normalized entropy (LN-Entropy)~\citep{malinin2020uncertainty} estimates uncertainty over multiple sampled responses.
Internal-state methods derive correctness scores from internal representations without training on correctness-labelled data. EigenScore~\citep{chen2024inside} measures consistency among the hidden representations of multiple generated responses, while CoE-R and CoE-C~\citep{wang2025coe} assess correctness from the magnitude and angular changes of hidden-state trajectories within a response.

\noindent\textbf{Evaluation Metrics.}
We adopt AUROC~\citep{boyd2013area}, FPR95, and AUPR~\citep{manning1999foundations} to evaluate how well each method distinguishes correct from incorrect responses.
We treat correct responses as the positive class.
AUROC measures the overall ranking quality between correct and incorrect responses across all decision thresholds.
FPR95 denotes the proportion of incorrect responses that are wrongly accepted as correct when the recall of correct responses reaches 95\%.
AUPR reflects the precision–recall trade-off for correct responses and remains informative under class imbalance.
For the mathematics and reasoning domains, we report the macro average over their two benchmarks so that each benchmark contributes equally.

\noindent\textbf{Hyperparameters.}
Our method has two fusion hyperparameters: $\alpha$ balances CIC and RG, while $\lambda$ controls the contribution of LC.
In addition, we set the maximum layer separation in CIC to $R=4$.
For each benchmark, we randomly split the questions into disjoint development and test sets with a ratio of 1:4 using a fixed random seed, and the same split is shared by all five models.
Only the development sets are used for hyperparameter selection, and all reported results are computed on the test sets.
Since the relative reliability of the signals varies across domains, we select one $(\alpha,\lambda)$ pair for each domain and share it across all five models and all three metrics.
% For each domain, we average the results of each model and metric over the development sets of its benchmarks, and select the pair that achieves the best performance on the largest number of model--metric entries.
The selected pairs are $(0.30, 0.07)$ for mathematics, $(0.07, 0.05)$ for reasoning, $(0.96, 0.21)$ for knowledge, and $(0.33, 0.06)$ for understanding.
Details of the search procedure are provided in Appendix~\ref{app:additional-experimental-results}.

\begin{figure}[!t]
  \centering
  \includegraphics[width=0.98\linewidth]
  {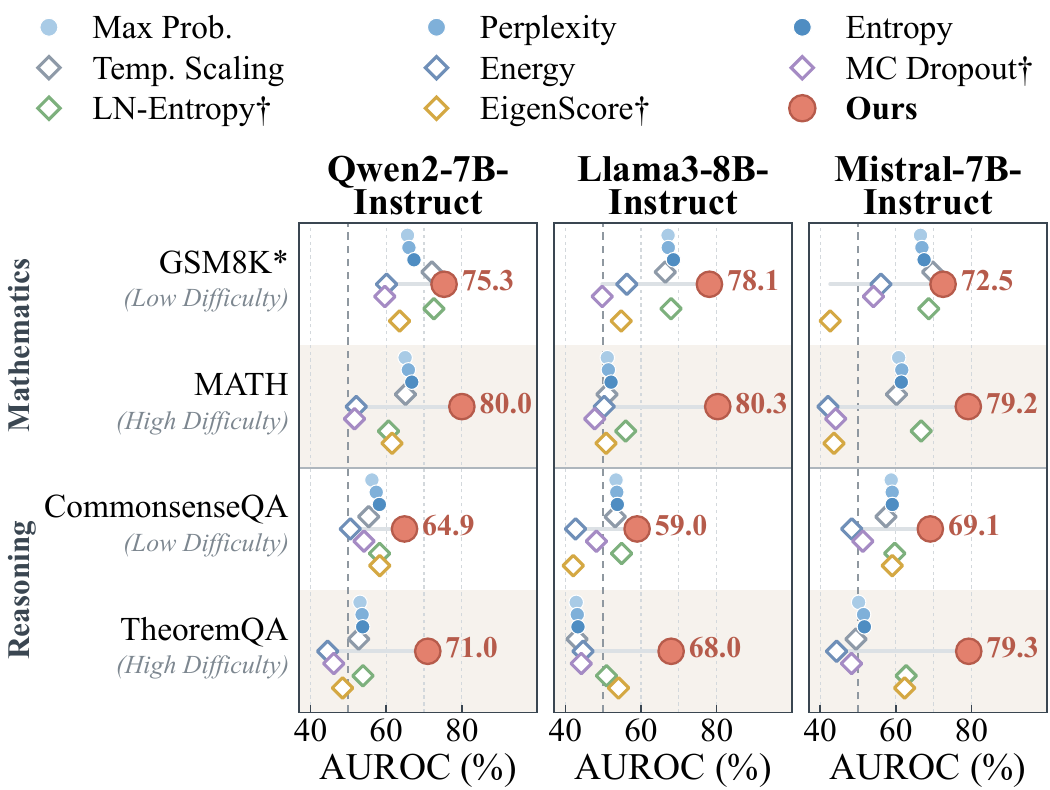}
  \vspace{-0.4cm}
\caption{\textbf{Self-evaluation across task difficulties.}
AUROC (\%) on mathematics and reasoning benchmarks of low and high difficulty with three models.
Our method (red circles) achieves the highest AUROC on all benchmarks, and its margin over the strongest baseline is larger on high-difficulty benchmarks.
Dashed lines indicate chance-level performance (50\%).
}
  \vspace{-0.6cm}
  \label{fig:task_difficulty}
\end{figure}

\subsection{Main Results}
Table~\ref{tab:circe-four-domains} reports the evaluation performance of our method and the baselines across five models and four domains.
Among the baselines, the strongest method varies across domains: CoE-C performs best in the mathematics domain, CoE-R in the reasoning and understanding domains, and LN-Entropy in the knowledge domain, indicating that no single baseline consistently reflects response correctness.
In contrast, our method achieves the best average AUROC, FPR95, and AUPR in all four domains, improving AUROC over the strongest baseline by 7.37\%, 3.13\%, 5.11\%, and 5.31\% in the mathematics, reasoning, knowledge, and understanding domains, respectively.
Notably, our method requires only a single generated response, whereas sampling-based baselines rely on multiple stochastic inferences.
However, the improvement is not uniform across settings, as our method achieves the highest AUROC in 18 of the 20 model–domain pairs, with the exceptions of Qwen2-7B-Instruct in the knowledge domain and Qwen2.5-14B-Instruct in the understanding domain, where it is slightly lower than the PSA pipeline and LN-Entropy, respectively.
Overall, these results suggest that combining internal computational signals with output confidence provides more reliable evidence for self-evaluation.
We further compare our method with supervised methods that train classifiers on the hidden states of LLMs in Appendix~\ref{app:source-trained}.

\subsection{Ablation Study}
To assess each component's contribution, we evaluate CIC, RG, and LC individually, their pairwise combinations, and the complete score. 
Table~\ref{tab:circe_ablation} reports the AUROC over six benchmarks for each model. 
The complete score achieves the best AUROC on all five models, with an average of 71.76\%. 
Removing any component consistently degrades performance, reducing the average AUROC by 2.05\%, 2.55\%, and 2.89\% without CIC, RG, and LC, respectively. Notably, the contribution of a component is not determined by its standalone performance. LC alone yields the lowest average AUROC of 59.52\%, yet removing it from the complete score causes a drop comparable to that of the other two components. Similarly, CIC and RG achieve comparable average AUROC when used alone, at 65.90\% and 66.16\% respectively, but their relative strength varies across models, with RG outperforming CIC on Qwen2-1.5B-Instruct and CIC outperforming RG by 6.68\% on Qwen2.5-14B-Instruct. Consistent with this complementarity, every pairwise combination outperforms each component on average. Overall, these results indicate that CIC, RG, and LC capture complementary correctness evidence, supporting their integration into a unified score.

\subsection{Extended Analysis}

\begin{figure}[!t]
  \centering
  \includegraphics[width=\linewidth]
  {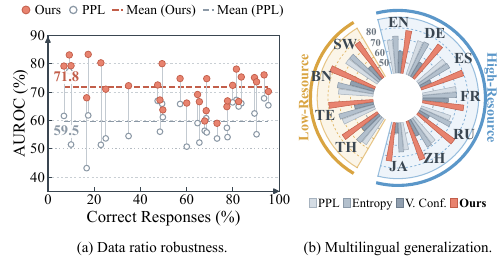}
  \vspace{-0.5cm}
  \caption{\textbf{Robustness to data ratio and generalization across languages.}
(a) AUROC (\%) of Circe and Perplexity on all 30 combinations of the five models and six benchmarks, plotted against the proportion of correct responses. Dashed lines indicate the mean AUROC.
(b) AUROC (\%) of our method and baselines on the 11 language versions of MGSM with Llama3-8B-Instruct.}
  \label{fig:data_ratio_robustness}
  \vspace{-0.5cm}
\end{figure}

\noindent\textbf{Effect of Task Difficulty.}
Tasks within the same domain can differ substantially in difficulty, which may affect the reliability of self-evaluation signals. We therefore compare a lower-difficulty and a higher-difficulty benchmark in each of the mathematics and reasoning domains, namely GSM8K and MATH for mathematics, and CommonsenseQA and TheoremQA for reasoning. As shown in Figure~\ref{fig:task_difficulty}, our method achieves the highest AUROC on all four benchmarks across the three models, and its margin over the strongest baseline is consistently larger on the higher-difficulty benchmark in each domain. On MATH, our method achieves an AUROC of 79.2\%--80.3\%.
On TheoremQA, our method achieves an AUROC of 68.0\%--79.3\%. Moreover, our method achieves a higher AUROC on the higher-difficulty benchmark in every model--domain pair, whereas most baselines show no such improvement. One possible explanation is that more difficult questions elicit more complex internal computation, so that attention--FFN interactions and representation changes reveal correctness-related patterns that output confidence alone fails to capture.

\noindent\textbf{Data Ratio Robustness.}
The proportion of correct responses depends on model accuracy and varies substantially across tasks, so a reliable correctness metric should remain discriminative regardless of whether correct responses form the majority or the minority. Figure~\ref{fig:data_ratio_robustness} (a) plots the AUROC of our method and perplexity against the proportion of correct responses for all 30 combinations of the five models and six benchmarks, where this proportion ranges from 6.9\% for Mistral-7B-Instruct on MATH to 95.3\% for Qwen2.5-14B-Instruct on Belebele.
Across this range, our method remains above the random level of 50\% in all combinations and shows no systematic decline toward either extreme, whereas perplexity falls below 50\% in several low-accuracy combinations. Overall, our method achieves higher average AUROC than perplexity across all four domains. The three exceptions are Qwen2-1.5B-Instruct on MMLU and Belebele, and Qwen2.5-14B-Instruct on CommonsenseQA, where the gap for the latter is only 1.3\%. These results indicate that our method is robust to varying class proportions, which is important in practice because the prevalence of incorrect responses changes with model capability and task difficulty.

\noindent\textbf{Multilingual Generalization.}
To assess the applicability of Circe in multilingual settings, we further examine its performance on MGSM, a multilingual benchmark released by \citet{shi2022language}, in which GSM8K problems are manually translated into ten typologically diverse languages, yielding 11 language versions together with English. Using Llama3-8B-Instruct, we evaluate each language version separately and compute AUROC within each language. We adopt the language grouping of \citet{shi2022language}, which divides the languages according to their frequency in pre-training data into seven high-resource languages, namely English (EN), German (DE), Spanish (ES), French (FR), Russian (RU), Chinese (ZH), and Japanese (JA), and four low-resource languages, namely Thai (TH), Telugu (TE), Bengali (BN), and Swahili (SW). As shown in Figure~\ref{fig:data_ratio_robustness} (b), our method achieves higher AUROC than perplexity, entropy, and verbal conf. in all 11 languages. Moreover, its advantage is not confined to high-resource languages and remains evident in the low-resource group. These results indicate that the correctness signals captured by our method generalize across languages, including those that are underrepresented in pre-training data.

\begin{table}[!t]
\centering
\caption{Multimodal generalization on Geometry3K with Qwen2.5-VL-7B. Our method outperforms baselines on AUROC (\%, $\uparrow$), AUPR (\%, $\uparrow$), and FPR95 (\%, $\downarrow$).}
\vspace{-0.3cm}
\label{tab:circe-e50-multimodal}
\begingroup
\small
\setlength{\tabcolsep}{3pt}
\renewcommand{\arraystretch}{1.12}
\begin{tabularx}{\columnwidth}{l*{3}{>{\raggedleft\arraybackslash}X}}
\toprule
\textbf{Method} & \textbf{AUROC} & \textbf{AUPR} & \textbf{FPR95} \\
\midrule
Max Prob.~\citep{hendrycks2017baseline} & 61.60 & 64.28 & 88.84 \\
Perplexity~\citep{Perplexity} & 61.91 & 64.80 & 88.84 \\
Entropy~\citep{Entropy} & 61.71 & 64.68 & 91.32 \\
CoE-R~\citep{wang2025coe} & 65.86 & 72.44 & 80.58 \\
CoE-C~\citep{wang2025coe} & 64.08 & 71.71 & 90.08 \\
\midrule
\rowcolor{domainblue}
\textbf{Ours} & \textbf{71.25} & \textbf{77.67} & \textbf{77.27} \\
\bottomrule
\end{tabularx}
\par\smallskip
\begin{minipage}{\columnwidth}
\end{minipage}
\endgroup
\vspace{-0.3cm}
\end{table}

\noindent\textbf{Multimodal Generalization.}
To evaluate whether our method generalizes to tasks beyond text, we further apply it to Geometry3K~\citep{lu2021inter}, a geometry problem-solving benchmark in which each problem is paired with a diagram, so that visual information is involved in response generation. Following the evaluation protocol of GeoR~\citep{bi2026geometry}, we use Qwen2.5-VL-7B~\citep{bai2025qwen25vl}, an open-source vision--language model, to generate responses. In addition to output-based baselines, we compare with CoE-R and CoE-C, which estimate correctness from hidden-state trajectories. As shown in Table~\ref{tab:circe-e50-multimodal}, our method achieves an AUROC of 71.25\%, exceeding the strongest baseline, CoE-R, by 5.39\%, and also achieves the highest AUPR and the lowest FPR95 among all methods. Notably, these results are obtained with the same fusion weights as in the text-only experiments and without additional training, indicating that our method transfers to visual--textual settings and captures correctness evidence beyond hidden-state trajectories alone.

\begin{table}[!t]
\centering
\caption{Robustness under open-world distribution shift. AUROC (\%, $\uparrow$) on mixed math and mixed reasoning with Llama3-8B-Instruct and Mistral-7B-Instruct. 
% Our method achieves the best AUROC in both settings.
}
\vspace{-0.3cm}
\label{tab:circe-mixed-distribution}
\begingroup
\small
\setlength{\tabcolsep}{3pt}
\renewcommand{\arraystretch}{1.12}
\begin{tabularx}{\columnwidth}{
  l
  >{\hsize=1.5\hsize
    \linewidth=\hsize
    \raggedright\arraybackslash}X
  *{2}{>{\hsize=0.75\hsize
         \linewidth=\hsize
         \centering\arraybackslash}X}
}
\toprule
\textbf{Model} & \textbf{Method}
& \shortstack{\textbf{Mixed}\\\textbf{Math}}
& \shortstack{\textbf{Mixed}\\\textbf{Reasoning}} \\
\midrule
\multirow{6}{*}{Llama3-8B-Instruct\citep{meta2024llama3}}
 & Verbal Conf.~\citep{kadavath2022} & 79.26 & 60.03 \\
 & Max Prob.~\citep{hendrycks2017baseline}       & 47.42 & 48.29 \\
 & Entropy~\citep{Entropy}         & 48.37 & 49.65 \\
 & CoE-R~\citep{wang2025coe}           & 78.47 & \underline{60.31} \\
 & CoE-C~\citep{wang2025coe}           & \underline{79.89} & 54.81 \\
\rowcolor{domainblue}
 & \textbf{Ours}   & \textbf{83.17} & \textbf{62.04} \\
\midrule
\multirow{6}{*}{Mistral-7B-Instruct\citep{jiang2023mistral}}
 & Verbal Conf.~\citep{kadavath2022} & 57.93 & 54.36 \\
 & Max Prob.~\citep{hendrycks2017baseline}       & 64.81 & 49.49 \\
 & Entropy~\citep{Entropy}         & 66.40 & 49.44 \\
 & CoE-R~\citep{wang2025coe}           & 62.77 & \underline{61.70} \\
 & CoE-C~\citep{wang2025coe}           & \underline{72.36} & 58.88 \\
\rowcolor{domainblue}
 & \textbf{Ours}   & \textbf{72.87} & \textbf{62.65} \\
\bottomrule
\end{tabularx}
\par\smallskip
\begin{minipage}{\columnwidth}
\footnotesize
\end{minipage}
\endgroup
\vspace{-0.5cm}
\end{table}

\noindent\textbf{Robustness under Open-World Distribution Shift.}
In practical deployment, questions often come from heterogeneous sources with different characteristics and difficulty levels. To evaluate whether our method remains effective under such distribution shifts, we construct two mixed settings that simulate open-world scenarios, with equal contributions from each source: mixed math, which combines GSM8K and TheoremQA, and mixed reasoning, which combines CommonsenseQA and MMLU. 
As shown in Table~\ref{tab:circe-mixed-distribution}, our method achieves the highest AUROC on both mixtures with both Llama3-8B-Instruct and Mistral-7B-Instruct. For instance, on mixed math with Llama3-8B-Instruct, our method achieves an AUROC of 83.17\%, surpassing the strongest baseline, CoE-C at 79.89\%, by 3.28\%. These results indicate that our method remains robust when responses from different sources are evaluated together.

\noindent\textbf{Computational Efficiency.} To evaluate the computational cost of our method, we report the mean execution time per response on GSM8K with Llama3-8B-Instruct. Table~\ref{tab:circe-efficiency-gsm8k} reports the costs of base LLM inference, softmax computation, and the additional computation of our method. All methods require base LLM inference to generate a response, while softmax computation converts output logits into probability distributions over the vocabulary. The latter is required by baselines such as perplexity and entropy. For our method, the reported additional computation covers all computations introduced by our method during and after response generation. It takes 0.67s per response, corresponding to approximately 18.0\% of the base inference time and giving a reconstructed total execution time of 4.36s per response. In contrast, methods based on sampling require multiple stochastic inferences, so their computational cost increases with the number of sampled responses. Overall, these results show that our method combines signals from internal states and model outputs while keeping the additional computational cost below that of base inference in this setting.

% 效率实验
\begin{table}[!t]
\centering
\caption{Computational efficiency on GSM8K with Llama3-8B-Instruct. Values are mean $\pm$ standard deviation. For our method, the reported time covers all additional computations during and after response generation.}
\vspace{-0.3cm}
\label{tab:circe-efficiency-gsm8k}
\normalsize
\setlength{\tabcolsep}{6pt}
\renewcommand{\arraystretch}{1.20}

\begin{tabularx}{\columnwidth}{>{\raggedright\arraybackslash}X>{\raggedleft\arraybackslash}X}
\toprule
Operation & Time (s) \\
\midrule
Base LLM inference & $3.69$\,\textcolor{black!55}{\footnotesize$\pm 1.27$} \\
Softmax computation & $0.07$\,\textcolor{black!55}{\footnotesize$\pm 0.21$} \\
\rowcolor{domainblue}
\textbf{Circe computation} & $\mathbf{0.67}$\,\textcolor{black!55}{\footnotesize$\pm 0.43$} \\
\bottomrule
\end{tabularx}
\vspace{-0.3cm}
\end{table}

\begin{figure}[!t]
\centering
\begin{minipage}[t]{0.48\linewidth}
  \centering
  \includegraphics[width=\linewidth]{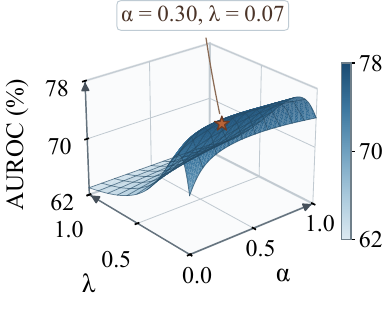}
  \par\vspace{-3pt}
  \small (a) Mathematics
\end{minipage}\hfill
\begin{minipage}[t]{0.48\linewidth}
  \centering
  \includegraphics[width=\linewidth]{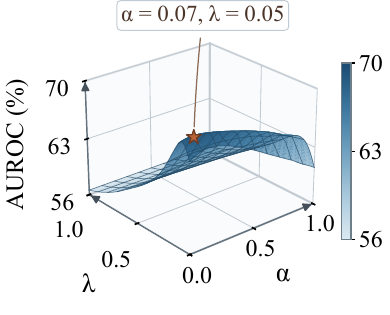}
  \par\vspace{-3pt}
  \small (b) Reasoning
\end{minipage}

\vspace{0.1em}

\begin{minipage}[t]{0.48\linewidth}
  \centering
  \includegraphics[width=\linewidth]{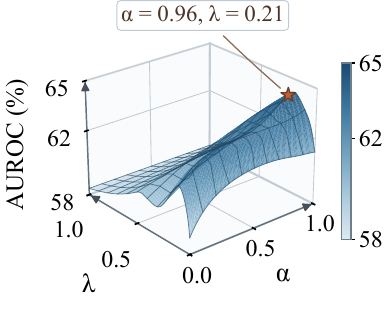}
  \par\vspace{-3pt}
  \small (c) Knowledge
\end{minipage}\hfill
\begin{minipage}[t]{0.48\linewidth}
  \centering
  \includegraphics[width=\linewidth]{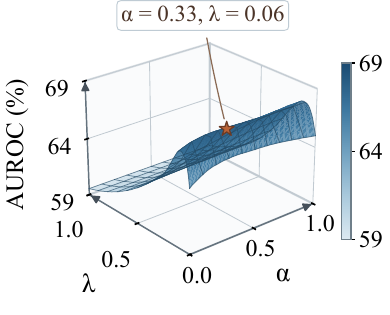}
  \par\vspace{-3pt}
  \small (d) Understanding
\end{minipage}
\caption{Hyperparameter sensitivity across four domains. Surface height and color indicate the AUROC (\%) of our method averaged over the benchmarks in each domain and the five models. Stars mark the selected setting in each domain. The AUROC range is set separately for each domain.}
\label{fig:circe-hyperparameter-sensitivity}
\vspace{-0.5cm}
\end{figure}

\noindent\textbf{Hyperparameter Sensitivity Across Domains.}
Figure~\ref{fig:circe-hyperparameter-sensitivity} shows the AUROC of our method over the search grid in each domain, averaged over the test sets of the benchmarks in the domain and the five models.
Overall, AUROC varies smoothly with the hyperparameters, and each domain contains a broad high-AUROC region rather than an isolated peak.
In the mathematics, reasoning, and understanding domains, AUROC remains close to its maximum over a wide range of $\alpha$ when $\lambda$ is small, so the results do not rely on a precise choice of either hyperparameter.
In the knowledge domain, AUROC increases steadily with $\alpha$ and peaks near $\alpha=1$, indicating that CIC is the more informative internal signal for knowledge questions.
In all four domains, AUROC drops markedly as $\lambda$ approaches 1, where the fused score reduces to LC alone, which confirms the benefit of combining internal signals with LC.

\section{Conclusion}
In this paper, we study how to evaluate the correctness of LLM responses without reference answers or repeated sampling. We observe that correct responses tend to exhibit higher cross-layer directional consistency between attention updates and the FFN output changes, revealing a component-level correctness signal that is not captured by output confidence or hidden-state trajectories alone. Building on this observation, we propose Circe, a self-evaluation method that assesses a response by combining three complementary signals: component interaction consistency, representation geometry, and likelihood confidence. Extensive experiments on five LLMs and six benchmarks spanning four domains show that our method achieves the best average performance across all evaluation metrics. Further analyses demonstrate that our method remains effective on more difficult tasks, under varying proportions of correct responses, under open-world distribution shift, and in multilingual and multimodal settings, while adding minimal scoring overhead.

% \clearpage

\bibliographystyle{ACM-Reference-Format}
\bibliography{www2027_conference}

% \clearpage

\appendix
\section*{Appendix}
\raggedbottom
\setlength{\abovedisplayskip}{4pt plus 2pt minus 2pt}
\setlength{\belowdisplayskip}{4pt plus 2pt minus 2pt}
\setlength{\abovedisplayshortskip}{2pt plus 1pt}
\setlength{\belowdisplayshortskip}{2pt plus 1pt}
\setlength{\textfloatsep}{11pt plus 2pt minus 2pt}
\setlength{\floatsep}{6pt plus 2pt minus 2pt}
Appendix~\ref{app:theory} gives a theoretical analysis of our method. It
explains what CIC measures and derives when fusing the scores improves AUROC.
Appendix~\ref{app:hyperparameters} describes how the fusion hyperparameters
are selected.
Appendix~\ref{app:source-trained} compares our method with label-based
methods that train classifiers on hidden states, and
Appendix~\ref{app:prompts} lists the instructions that all models follow
when generating responses to the benchmark questions.
Appendix~\ref{app:dis} discusses whether CIC captures component interactions
rather than vector similarity, why our method remains training-free despite
using development data, and what CIC contributes beyond representation
geometry and likelihood.

\vspace{-3pt}
\section{Theoretical Analysis}
\label{app:theory}

This section analyzes the three scores of our method and their fusion.
Appendix~\ref{app:theory-cic} shows that CIC measures whether later FFNs
respond to attention updates in directions consistent with earlier attention
updates. Appendix~\ref{app:theory-fusion} derives the condition under which
fusion improves AUROC, which applies both to the internal score
$s_{\mathrm{int}}$ that fuses CIC and RG and to the final score
$s_{\mathrm{Circe}}$ that fuses $s_{\mathrm{int}}$ with LC.
Throughout, $\|\cdot\|$ denotes the Euclidean norm and
$\cos(u,v)=u^{\top}v/(\|u\|\,\|v\|)$ for nonzero $u,v$.

\vspace{-3pt}
\subsection{What CIC Measures}
\label{app:theory-cic}

We assume that each FFN map $F_\ell$, including its input normalization, is
twice continuously differentiable on the segments
$\{h_{t,\ell-1}+\tau a_{t,\ell}:\tau\in[0,1]\}$, and that the operator norm of
its second derivative is bounded by $\beta_\ell$ on these segments. This holds
for smooth activations such as GELU and SiLU and for normalization layers away
from the zero vector. Let $J_{t,\ell}$ denote the Jacobian of $F_\ell$ at
$h_{t,\ell-1}$, so that $J_{t,\ell}\,a_{t,\ell}$ is the linear response of the
FFN to the attention update.

\begin{lemma}[Linearization of the attention-induced FFN output change]
\label{lem:cic-taylor}
For every token $t$ and block $\ell$,
\begin{equation}
    \Delta m_{t,\ell} = J_{t,\ell}\,a_{t,\ell} + e_{t,\ell},
    \qquad
    \|e_{t,\ell}\| \le \frac{\beta_\ell}{2}\,\|a_{t,\ell}\|^{2}.
\end{equation}
Consequently,
$\overline{\Delta m}_{\ell}=v_\ell+\bar e_\ell$ with
$v_\ell=\frac{1}{T}\sum_{t=1}^{T}J_{t,\ell}\,a_{t,\ell}$ and
$\|\bar e_\ell\|\le \frac{\beta_\ell}{2T}\sum_{t=1}^{T}\|a_{t,\ell}\|^{2}$.
\end{lemma}

\begin{proof}
Write $h=h_{t,\ell-1}$ and $a=a_{t,\ell}$. By Taylor's theorem with integral
remainder, $F_\ell(h+a)-F_\ell(h)=J_{t,\ell}\,a+\int_0^1(1-\tau)\,
D^{2}F_\ell(h+\tau a)[a,a]\,d\tau$, and the norm of the integral is at most
$\beta_\ell\|a\|^{2}\int_0^1(1-\tau)\,d\tau=\beta_\ell\|a\|^{2}/2$.
Averaging over $t$ and applying the triangle inequality gives the second
statement.
\end{proof}

\begin{lemma}[Stability of cosine similarity]
\label{lem:cos-perturb}
For nonzero $u,v\in\mathbb{R}^{d}$ and any $e$ with $\|e\|<\|v\|$,
\begin{equation}
    \bigl|\cos(u,v+e)-\cos(u,v)\bigr| \le \frac{2\|e\|}{\|v\|}.
\end{equation}
\end{lemma}

\begin{proof}
Let $\hat x=x/\|x\|$. By the Cauchy--Schwarz inequality,
$|\cos(u,v+e)-\cos(u,v)|=|\hat u^{\top}(\widehat{v+e}-\hat v)|
\le\|\widehat{v+e}-\hat v\|$. Moreover,
\[
\begin{aligned}
\|\widehat{v+e}-\hat v\|
&\le \frac{\|(v+e)-v\|}{\|v\|}
   +\|v+e\|\,\Bigl|\frac{1}{\|v+e\|}-\frac{1}{\|v\|}\Bigr|\\
&= \frac{\|e\|+\bigl|\|v\|-\|v+e\|\bigr|}{\|v\|}
\le \frac{2\|e\|}{\|v\|}.
\end{aligned}
\]
\end{proof}

\begin{proposition}[CIC as Jacobian-mediated alignment]
\label{prop:cic}
Let $\tilde s_{\mathrm{CIC}}$ be defined as $s_{\mathrm{CIC}}$ with
$\overline{\Delta m}_{\ell+r}$ replaced by
$v_{\ell+r}=\frac{1}{T}\sum_{t}J_{t,\ell+r}\,a_{t,\ell+r}$. If
$\|\bar e_{\ell}\|<\|v_{\ell}\|$ for every block $\ell$ that enters the score,
then
\begin{equation}
    \bigl|s_{\mathrm{CIC}}-\tilde s_{\mathrm{CIC}}\bigr|
    \le \max_{\ell}\frac{2\|\bar e_{\ell}\|}{\|v_{\ell}\|}
    \le \max_{\ell}\frac{\beta_\ell\,\frac{1}{T}\sum_{t}\|a_{t,\ell}\|^{2}}
                       {\|v_{\ell}\|}.
\end{equation}
\end{proposition}

\begin{proof}
$s_{\mathrm{CIC}}$ and $\tilde s_{\mathrm{CIC}}$ are the same convex
combination of cosine terms, so their difference is bounded by the largest
difference between corresponding terms. Lemma~\ref{lem:cos-perturb} with
$u=\bar a_\ell$, $v=v_{\ell+r}$, and $e=\bar e_{\ell+r}$ bounds each term, and
Lemma~\ref{lem:cic-taylor} bounds $\|\bar e_{\ell+r}\|$.
\end{proof}

\noindent\textbf{Interpretation.}
When the FFN maps are nearly linear over the range of the attention updates,
that is, when the right-hand side of the bound above is small, CIC is close to
$\tilde s_{\mathrm{CIC}}$. This approximation compares the response-level
attention update of block $\ell$ with the attention update of block $\ell+r$
after it is mapped through the local Jacobians of the FFN in that block.
A high CIC score therefore indicates that later FFNs turn their attention
input into output changes whose directions agree with earlier attention
updates, which is the cross-layer consistency that CIC is designed to
capture.

\vspace{-3pt}
\subsection{Justification for Score Fusion}
\label{app:theory-fusion}

Since our method fuses its scores with nonnegative weights, we analyze linear
fusion under a Gaussian model with a common covariance. Let
$S\in\mathbb{R}^{k}$ collect $k$ scores and assume
$S\mid Y=y\sim\mathcal{N}(\mu_y,\Sigma)$ for $y\in\{0,1\}$, with
$\Sigma\succ 0$, $Y=1$ for correct responses, and
$\Delta=\mu_1-\mu_0$. Although this model simplifies the actual score
distributions, it makes explicit how the correlation between scores
determines the benefit of fusion.

\begin{lemma}[AUROC of a linear score]
\label{lem:auroc-gauss}
For any $w\neq 0$, the AUROC of $w^{\top}S$ is
$\Phi\bigl(w^{\top}\Delta/\sqrt{2\,w^{\top}\Sigma w}\bigr)$, where $\Phi$ is
the cumulative distribution function of the standard normal distribution. It is maximized by
$w^{\star}=c\,\Sigma^{-1}\Delta$ for any $c>0$, with value
$\Phi\bigl(\sqrt{\Delta^{\top}\Sigma^{-1}\Delta/2}\bigr)$.
\end{lemma}

\begin{proof}
For independent $S_1$ and $S_0$ drawn from the two classes,
$w^{\top}(S_1-S_0)\sim\mathcal{N}(w^{\top}\Delta,\,2w^{\top}\Sigma w)$, so
$\Pr(w^{\top}S_1>w^{\top}S_0)=\Phi\bigl(w^{\top}\Delta/
\sqrt{2w^{\top}\Sigma w}\bigr)$. By the Cauchy--Schwarz inequality in the
inner product induced by $\Sigma$,
$(w^{\top}\Delta)^{2}\le(w^{\top}\Sigma w)(\Delta^{\top}\Sigma^{-1}\Delta)$,
with equality if and only if $w\propto\Sigma^{-1}\Delta$.
\end{proof}

Our method fuses its scores in two steps. The internal score
$s_{\mathrm{int}}$ combines CIC and RG, and the final score
$s_{\mathrm{Circe}}$ combines $s_{\mathrm{int}}$ with LC, so each step fuses
a pair of scores. For such a pair, orient each score so that correct responses score higher on
average, and standardize it by its within-class standard deviation, which
does not change its AUROC. Let $\delta_1\ge\delta_2>0$ be the resulting
standardized mean differences, so that score~1 is the stronger one, and let
$\rho\in(-1,1)$ be their within-class correlation. Since the overall scale of
$w$ does not affect AUROC, nonnegative fusion amounts to $w=(1,\gamma)$ with
$\gamma\ge 0$.

\begin{proposition}[Condition for a fusion gain]
\label{prop:fusion}
Under the Gaussian model, the following hold.
\begin{enumerate}[leftmargin=*,itemsep=0pt]
\item Some $\gamma>0$ yields a higher AUROC than the stronger score alone if
and only if $\rho<\delta_2/\delta_1$.
\item If $\rho<\delta_2/\delta_1$, the best nonnegative fusion attains AUROC
$\Phi(D/\sqrt{2})$ with
\begin{equation}
    D^{2}
    = \delta_1^{2}+\frac{(\delta_2-\rho\,\delta_1)^{2}}{1-\rho^{2}}.
\end{equation}
\end{enumerate}
\end{proposition}

\begin{proof}
By Lemma~\ref{lem:auroc-gauss}, the AUROC of $S_1+\gamma S_2$ is increasing in
$g(\gamma)=(\delta_1+\gamma\delta_2)/\sqrt{1+2\gamma\rho+\gamma^{2}}$, and
$g(0)=\delta_1$. For $\gamma>0$,
$g(\gamma)>\delta_1$ is equivalent to
$(\delta_1+\gamma\delta_2)^{2}>\delta_1^{2}(1+2\gamma\rho+\gamma^{2})$, that
is, to
$2\delta_1(\delta_2-\rho\delta_1)+\gamma(\delta_2^{2}-\delta_1^{2})>0$. If
$\rho\ge\delta_2/\delta_1$, both terms are nonpositive because
$\delta_2\le\delta_1$, so no $\gamma>0$ improves on the stronger score. If
$\rho<\delta_2/\delta_1$, the inequality holds for all sufficiently small
$\gamma>0$, which proves item~1. For item~2,
$w^{\star}\propto\Sigma^{-1}\Delta=\frac{1}{1-\rho^{2}}
(\delta_1-\rho\delta_2,\ \delta_2-\rho\delta_1)$, whose components are both
positive when $\rho<\delta_2/\delta_1$, so the unconstrained optimum is
feasible. Its separation is
$\Delta^{\top}\Sigma^{-1}\Delta
=\frac{\delta_1^{2}-2\rho\delta_1\delta_2+\delta_2^{2}}{1-\rho^{2}}
=\delta_1^{2}+\frac{(\delta_2-\rho\delta_1)^{2}}{1-\rho^{2}}$.
\end{proof}

\noindent\textbf{Complementarity of the Three Scores.}
Proposition~\ref{prop:fusion} makes the role of complementarity explicit.
A weaker score improves a stronger one only when their within-class
correlation is below the ratio $\delta_2/\delta_1$ of their separations, and
the attainable gain grows with $(\delta_2-\rho\delta_1)^{2}/(1-\rho^{2})$.
A weak score can therefore still be useful if it is weakly correlated with
the stronger one, whereas a strong score that is highly correlated with it
adds little. The ablation in Table~\ref{tab:circe_ablation} is consistent
with this analysis. LC has the lowest average AUROC among the single scores
at 59.52\%, yet fusing it with the internal score $s_{\mathrm{int}}$ raises the
average AUROC from 68.87\% to 71.76\%. Under the Gaussian model, such a gain requires
the likelihood signal to carry information about correctness that is not
already captured by the internal signals CIC and RG.

\vspace{-3pt}
\section{Hyperparameter Selection and Sensitivity}
\label{app:hyperparameters}
\label{app:additional-experimental-results}

Circe involves two groups of hyperparameters: the fusion hyperparameters $\alpha$ and $\lambda$, which weight the three component scores, and the maximum layer separation $R$, which determines the layer pairs used in CIC. We describe how each is set below.

\vspace{-3pt}
\subsection{Fusion Weights}
\label{app:fusion-weights}

\noindent\textbf{Selection Procedure.}
The two fusion hyperparameters $\alpha$ and $\lambda$ assign the weights $w_{\mathrm{CIC}}=(1-\lambda)\alpha$, $w_{\mathrm{RG}}=(1-\lambda)(1-\alpha)$, and $w_{\mathrm{LC}}=\lambda$ to CIC, RG, and LC, respectively, which sum to one.
We search all such weight vectors on a grid with a step size of 0.01, namely $\{(i/100,\,j/100,\,k/100) \mid i,j,k\in\mathbb{Z}_{\geq 0},\ i+j+k=100\}$, which contains 5,151 combinations.
All selection is performed on the development sets, which are obtained by randomly splitting the questions of each benchmark into development and test sets with a ratio of 1:4.
For each domain and each candidate weight, we compute 15 results, one for each of the five models under each of the three metrics, where each result is averaged over the development sets of the benchmarks in the domain.
We compare each result with the strongest baseline under the same model and metric, and select the weight that outperforms the strongest baseline in the largest number of the 15 results.
The selected weight vector is shared across all five models and all three metrics within the domain and is fixed for all test-set evaluations.
The selected weights $(w_{\mathrm{CIC}}, w_{\mathrm{RG}}, w_{\mathrm{LC}})$ are $(0.28, 0.65, 0.07)$ for mathematics, $(0.07, 0.88, 0.05)$ for reasoning, $(0.76, 0.03, 0.21)$ for knowledge, and $(0.31, 0.63, 0.06)$ for understanding, corresponding to $(\alpha,\lambda)\approx(0.30, 0.07)$, $(0.07, 0.05)$, $(0.96, 0.21)$, and $(0.33, 0.06)$, respectively.

\vspace{-3pt}
\subsection{Maximum Layer Separation}
\label{app:layer-lag}

The maximum layer separation $R$ controls the range of layer
separations included in CIC, and we fix $R=4$ in all experiments.
Table~\ref{tab:circe-R-domain} examines this choice across the four domains with Qwen2-7B-Instruct.
At $R=4$, AUROC remains close to the best displayed result in every domain.
Increasing $R$ further produces only small, domain-dependent changes rather than a consistent improvement, while the number of layer-pair terms in the score grows with $R$.
$R=4$ therefore provides a compact multi-lag setting without sacrificing detection performance.

\begin{table}[!t]
\centering
\caption{\textbf{Sensitivity to the maximum layer separation.}
AUROC (\%, $\uparrow$) on Qwen2-7B-Instruct, averaged over the
benchmarks in each domain.}
\vspace{-0.4cm}
\label{tab:circe-R-domain}
\begingroup
\small
\setlength{\tabcolsep}{3pt}
\renewcommand{\arraystretch}{1.12}
\begin{tabularx}{\columnwidth}{@{}l>{\raggedleft\arraybackslash}X>{\columncolor{domainblue}\raggedleft\arraybackslash}X*{3}{>{\raggedleft\arraybackslash}X}@{}}
\toprule
\textbf{Domain} & \textbf{\boldmath$R=1$} & \textbf{\boldmath$R=4$} & \textbf{\boldmath$R=8$} & \textbf{\boldmath$R=16$} & \textbf{\boldmath$R=27$} \\
\midrule
Mathematics & 77.37 & \textbf{77.68} & 77.65 & 77.64 & 77.67 \\
Reasoning & 67.93 & \textbf{67.94} & 67.93 & 67.87 & 67.89 \\
Knowledge & 59.60 & 59.79 & \textbf{59.81} & 59.74 & 59.62 \\
Understanding & 74.85 & \textbf{75.10} & 75.06 & 74.93 & 74.72 \\
\bottomrule
\end{tabularx}
\endgroup
% \vspace{-0.1cm}
\end{table}

\vspace{-3pt}
\section{Label-Based Evaluation with Hidden States}
\label{app:source-trained}

ITI~\citep{li2023iti} and MIND~\citep{su2024mind} are supervised methods that train classifiers on LLM hidden states.
ITI trains linear probes on the labelled candidate answers of TruthfulQA~\citep{lin2022truthfulqa}, and MIND trains a classifier on hallucination labels automatically constructed from Wikipedia.
Since these methods also rely on internal states, we compare them with our method on TruthfulQA, GSM8K, MATH, and MMLU with Llama3-8B-Instruct.
On all four benchmarks, we evaluate the correctness of responses generated by the LLM itself rather than the labelled candidate answers, so the evaluated responses differ from the training data of both methods in task type, data source, or the distribution of responses.

As shown in Table~\ref{tab:circe-e48-table4-mmlu}, the supervised methods ITI and MIND remain close to chance on all four benchmarks, with AUROC between 45.83\% and 57.51\%.
Even on TruthfulQA, whose questions are used to train ITI, its probe does not transfer from labelled candidate answers to LLM-generated responses, suggesting that supervised classifiers capture features specific to their training data.
In contrast, all three label-free methods exceed 55\% AUROC on every benchmark, and our method achieves the highest AUROC on all four benchmarks.
These results indicate that fitting a classifier with labels does not guarantee reliable correctness evaluation on new tasks, whereas label-free signals derived from internal states transfer more consistently across different task types.

\begin{table}[!t]
\centering
\caption{\textbf{Comparison with label-based methods.} AUROC (\%, $\uparrow$) of the label-based methods ITI and MIND and the label-free methods CoE-R, CoE-C, and ours, evaluated on responses generated by Llama3-8B-Instruct.}
\vspace{-0.3cm}
\label{tab:circe-e48-table4-mmlu}
\begingroup
\small
\setlength{\tabcolsep}{3pt}
\renewcommand{\arraystretch}{1.12}
\begin{tabularx}{\columnwidth}{>{\raggedright\arraybackslash}X*{4}{>{\raggedleft\arraybackslash}X}}
\toprule
\textbf{Method} & \textbf{TruthfulQA} & \textbf{GSM8K} & \textbf{MATH} & \textbf{MMLU} \\
\midrule
ITI~\citep{li2023iti} & 48.06 & 48.36 & 57.51 & 53.09 \\
MIND~\citep{su2024mind} & 51.95 & 53.03 & 56.89 & 45.83 \\
CoE-R~\citep{wang2025coe} & 57.15 & 71.15 & 75.23 & 60.19 \\
CoE-C~\citep{wang2025coe} & 60.92 & 73.11 & 74.97 & 55.67 \\
\midrule
\rowcolor{domainblue}
\textbf{Ours} & \textbf{62.85} & \textbf{76.14} & \textbf{79.84} & \textbf{66.31} \\
\bottomrule
\end{tabularx}
\endgroup
\end{table}

\vspace{-3pt}
\section{Response Generation Instructions}
\label{app:prompts}

All models generate responses with their official chat templates and the
dataset-specific instructions listed below. Each instruction describes the
task, asks for step-by-step reasoning, and fixes the format of the final
answer so that the answer can be extracted and compared with the reference.
The placeholder \texttt{\{input\_data\}} is replaced by the test input, and
\texttt{\{answer\_type\}} is replaced by the answer type provided by
TheoremQA.

% GSM8K & MGSM
\begin{tcolorbox}[
  width=\columnwidth,
  colframe=promptnavy,
  colback=promptbackground,
  colbacktitle=promptnavy,
  coltitle=black,
  title={GSM8K and MGSM},
  fonttitle=\small\bfseries,
  fontupper=\footnotesize,
  boxrule=0.7pt,
  arc=2mm,
  boxsep=0pt,
  left=6pt,
  right=6pt,
  top=3pt,
  bottom=3pt,
  toptitle=1.5pt,
  bottomtitle=1.5pt,
  before skip=3pt,
  after skip=3pt
]
Solve this math problem. Give the reasoning steps before giving the final answer on the last line by itself in the format of \texttt{"Answer:"}. Do not add anything other than the integer answer after \texttt{"Answer:"}.

\textbf{Question:} \texttt{\{input\_data\}}
\end{tcolorbox}

% MATH 
\begin{tcolorbox}[
  width=\columnwidth,
  colframe=promptnavy,
  colback=promptbackground,
  colbacktitle=promptnavy,
  coltitle=black,
  title={MATH},
  fonttitle=\small\bfseries,
  fontupper=\footnotesize,
  boxrule=0.7pt,
  arc=2mm,
  boxsep=0pt,
  left=6pt,
  right=6pt,
  top=3pt,
  bottom=3pt,
  toptitle=1.5pt,
  bottomtitle=1.5pt,
  before skip=3pt,
  after skip=3pt
]
Please reason step by step, and put your final answer within \texttt{\textbackslash boxed\{\{\}\}}\par
\textbf{Question:} \texttt{\{input\_data\}}\par
\end{tcolorbox}

% CommonsenseQA
\begin{tcolorbox}[
  width=\columnwidth,
  colframe=promptnavy,
  colback=promptbackground,
  colbacktitle=promptnavy,
  coltitle=black,
  title={CommonsenseQA},
  fonttitle=\small\bfseries,
  fontupper=\footnotesize,
  boxrule=0.7pt,
  arc=2mm,
  boxsep=0pt,
  left=6pt,
  right=6pt,
  top=3pt,
  bottom=3pt,
  toptitle=1.5pt,
  bottomtitle=1.5pt,
  before skip=3pt,
  after skip=3pt
]
Answer the following multiple choice common-sense reasoning question. The last line of your response should be of the following format: \texttt{'Answer: \$LETTER'} (without quotes) where LETTER is one of ABCDE. Think step by step and output the reasoning process before answering.

\textbf{Question:} \texttt{\{input\_data\}}
\end{tcolorbox}

% TheoremQA 
\begin{tcolorbox}[
  width=\columnwidth,
  colframe=promptnavy,
  colback=promptbackground,
  colbacktitle=promptnavy,
  coltitle=black,
  title={TheoremQA},
  fonttitle=\small\bfseries,
  fontupper=\footnotesize,
  boxrule=0.7pt,
  arc=2mm,
  boxsep=0pt,
  left=6pt,
  right=6pt,
  top=3pt,
  bottom=3pt,
  toptitle=1.5pt,
  bottomtitle=1.5pt,
  before skip=3pt,
  after skip=3pt
]
Below is an instruction that describes a task, paired with an input that provides further context. Write a response that appropriately completes the request.

\textbf{Instruction:}\par
Please read a math problem, and then think step by step to derive the answer. The answer is decided by Answer Type.\par
If the Answer type in [bool], the answer needs to be True or False.\par
Else if the Answer type in [integer, float] , The answer needs to be in numerical form.\par
Else if the Answer type in [list of integer, list of float] , the answer needs to be a list of number like [2, 3, 4].\par
Else if the Answer type in [option], the answer needs to be an option like (a), (b), (c), (d).\par
You need to output the answer in your final sentence like 'Therefore, the answer is ...'.

\textbf{Question:} \texttt{\{input\_data\}}

\textbf{Answer\_type:} \texttt{\{answer\_type\}}

\end{tcolorbox}

% MMLU
\begin{tcolorbox}[
  width=\columnwidth,
  colframe=promptnavy,
  colback=promptbackground,
  colbacktitle=promptnavy,
  coltitle=black,
  title={MMLU},
  fonttitle=\small\bfseries,
  fontupper=\footnotesize,
  boxrule=0.7pt,
  arc=2mm,
  boxsep=0pt,
  left=6pt,
  right=6pt,
  top=3pt,
  bottom=3pt,
  toptitle=1.5pt,
  bottomtitle=1.5pt,
  before skip=3pt,
  after skip=3pt
]
Answer the following multiple choice question. The last line of your response should be of the following format: \texttt{'Answer: \$LETTER'} (without quotes) where LETTER is one of ABCD. Think step by step before answering.

\textbf{Question:} \texttt{\{input\_data\}}
\end{tcolorbox}

% Belebele
\begin{tcolorbox}[
  width=\columnwidth,
  colframe=promptnavy,
  colback=promptbackground,
  colbacktitle=promptnavy,
  coltitle=black,
  title={Belebele},
  fonttitle=\small\bfseries,
  fontupper=\footnotesize,
  boxrule=0.7pt,
  arc=2mm,
  boxsep=0pt,
  left=6pt,
  right=6pt,
  top=3pt,
  bottom=3pt,
  toptitle=1.5pt,
  bottomtitle=1.5pt,
  before skip=3pt,
  after skip=0pt
]
Answer the following multiple choice reading-comprehension question. The last line of your response should be of the following format: \texttt{'Answer: \$LETTER'} (without quotes) where LETTER is one of ABCD. Please fully understand the passage and give explanations step by step before answering.

\textbf{Question:} \texttt{\{input\_data\}}
\end{tcolorbox}

\vspace{-4pt}
\section{More Discussions}
\label{app:dis}
\noindent $\triangleright$ \textbf{\textit{Does CIC characterize component interactions or merely vector similarity?}}\\
CIC characterizes component interactions. It does not compare earlier attention updates with raw FFN outputs, but with attention-induced FFN changes. Specifically, Eq.~(1) evaluates the same FFN with and without its current attention update while holding the incoming residual state and model parameters fixed, which isolates how attention modifies the FFN contribution within each block. CIC then measures whether these interaction-derived changes in later blocks align with earlier attention directions. Appendix~A further shows that, under a local linear approximation, these changes correspond to attention updates transformed by the FFN Jacobian. The compared vectors therefore have a specific computational interpretation, and their alignment reflects cross-layer consistency in attention--FFN processing.

\noindent $\triangleright$ \textbf{\textit{Why is Circe training-free despite using development data?}}\\
Training-free refers to the absence of parameter optimization: Circe neither updates the backbone nor trains an additional correctness detector, and its component scores are computed directly from model activations and token probabilities. Correctness-labelled development data are used only to select the fusion weights, with a single configuration shared across all five models and all three metrics within each domain. Development and test questions are disjoint, and the selected weights remain fixed during test evaluation. Thus, the scoring itself requires no training, and labelled data enter only through the selection of a few hyperparameters.

\noindent $\triangleright$ \textbf{\textit{What does CIC contribute beyond representation geometry and likelihood?}}\\
CIC contributes component-level evidence that neither RG nor LC captures. RG summarizes the overall hidden-state trajectory and LC measures output likelihood, but neither separates how attention changes FFN processing. Table~2 confirms this incremental contribution: adding CIC to RG and LC raises the average AUROC from 69.71\% to 71.76\%, with improvements on all five models. Moreover, although CIC and RG perform comparably alone, combining them outperforms either one for every model. These results indicate that CIC provides correctness evidence complementary to global representation geometry and output confidence.

\end{document}